\documentclass[electronics,article,accept,pdftex,moreauthors]{Definitions/mdpi} 

\firstpage{1} 
\pubvolume{13}
\issuenum{15}
\articlenumber{3091}
\pubyear{2024}
\copyrightyear{2024}
\datereceived{22 June 2024} 
\daterevised{28 July 2024} 
\dateaccepted{31 July 2024} 
\datepublished{5 August 2024} 
\hreflink{doi.org/10.3390/electronics13153091} 

\usepackage{xcolor}
\usepackage{algorithm,algpseudocode}
\usepackage{subcaption}

\newcommand{\E}{\mathbb{E}}

\Title{Illumination and Shadows in Head Rotation: experiments with Denoising Diffusion Models}

\TitleCitation{Illumination and Shadows in Head Rotation: experiments with Denoising Diffusion Models}

\Author{Andrea Asperti\orcidA{}, Gabriele Colasuonno and Antonio Guerra}

\AuthorNames{Andrea Asperti\orcidA{}, Gabriele Colasuonno and Antonio Guerra}

\AuthorCitation{Asperti, A.; Colasuonno, G.; Guerra A.}

\address{%
University of Bologna, Department of Informatics - Science and Engineering (DISI); andrea.asperti@unibo.it, salvatore.fiorilla@unibo.it, lorenzo.orisini4@studio.unibo.it
}

\corres{Correspondence: andrea.asperti@unibo.it}

\abstract{Accurately modeling the effects of illumination and shadows during head rotation is critical in computer vision for enhancing image realism and reducing artifacts. This study delves into the latent space of denoising diffusion models to identify compelling trajectories that can express continuous head rotation under varying lighting conditions. A key contribution of our work is the generation of additional labels from the CelebA dataset, categorizing images into three groups based on prevalent illumination direction: left, center, and right. These labels play a crucial role in our approach, enabling more precise manipulations and improved handling of lighting variations.
Leveraging a recent embedding technique for Denoising Diffusion Implicit Models (DDIM), our method achieves noteworthy manipulations, encompassing a wide rotation angle of $\pm 30$ degrees, while preserving individual distinct characteristics even under challenging illumination conditions. Our methodology involves computing trajectories that approximate clouds of latent representations of dataset samples with different yaw rotations through linear regression. Specific trajectories are obtained by analyzing subsets of data that share significant attributes with the source image, including light direction.
Notably, our approach does not require any specific training of the generative model for the task of rotation; we merely compute and follow specific trajectories in the latent space of a pre-trained face generation model. This article showcases the potential of our approach and its current limitations through a qualitative discussion of notable examples. This study contributes to the ongoing advancements in representation learning and the semantic investigation of the latent space of generative models.
}


\begin{document}



\section{Introduction}
The possibility of manipulating images acting on their latent representation, typical of generative models, has always exerted a particular fascination. 
Understanding the effect of a tiny modification to the encoding of the generated sample help us to better understand the properties of the latent space, 
and the disentanglement of the different features. This is strictly related 
to editing, since understanding semantically meaningful directions (such as color, pose, and shape) can be utilized to modify an image to include the desired features. 

The field of deep generative modeling has recently witnessed a significant shift with the emergence of Denoising Diffusion Models (DDM) \cite{DDPM}, which are rapidly establishing themselves as the new state-of-the-art technology \cite{conditioning_denoising,stable-diffusion}. These models are likely poised to surpass the long-standing dominance of Generative Adversarial Networks (GANs) \cite{Diff_vs_GAN}, joining an excellent generative quality with high sample diversity, simple and stable training and a 
solid probabilistic foundation. They achieved impressive results in a wide range of diversified domains comprising e.g. medical imaging \cite{PLOS2024}, healthcare \cite{Healthcare_review}, protein synthesis \cite{TrippeYTBBBJ23}
or weather forecasting \cite{ZhaoDWH24}. 

While DDMs have shown remarkable capabilities in generating realistic samples, the exploration of the latent space and the manipulation of generated samples to edit specific attributes remains a complex task. This is partly due to the high dimensionality of the latent space, which poses challenges in navigating and understanding the underlying semantics, but also to the complexity of embedding
data into the latent space, computing the internal encoding of a given sample. In the largely explored case of GANs,
most of the known techniques for semantic editing \cite{shen20,GANSpace,InterpretingLatent,shen22} 
are in fact based on the preliminary definition of a "recoder" \cite{gan_inversion_optimization,alaluf2022hyperstyle,gan_inversion_survey}, inverting the generative process, and essentially providing a functionality similar to encoders for Variational Autoencoders \cite{VAEKingma,VAEGreen}. 

The embedding problem for the particular but important case of Denoising Diffusion Implicit Models \cite{DDIM} has been recently investigated
in \cite{asperti2022embedding}. A crucial difference in the latent space of denoising models is that it appears to be organized as a foliation, with a different slice for each data point. These slices correspond to the set of all noisy points in the space that will collapse onto the given data point during the denoising process. Slices are typically very large, occupying significant portions of the input space. As a result, the embedding problem is inherently multimodal and underconstrained. Embedding techniques, such as the one described in \cite{asperti2022embedding}, typically select a point in the slice based on criteria that are difficult to control and decipher. Consequently, there is no evidence that we can organize the latent points extracted from the embedding network along meaningful trajectories. This is precisely the problem we aim to address in this work.

Unlike many works in the literature that focus on one-step modifications of the input (e.g., changing a color, adding or removing elements), we are interested in continuous modifications of the input image. The case of head rotation is particularly appealing for our study for several reasons. Firstly, face generation is a well-investigated domain, 
and the rotation problem is recognized as one of the most complex and intriguing editing operations.
The challenge with head rotation is that it requires preserving the distinctive features of the person while applying significant transformations that cannot be defined in terms of texture, color, shapes, or other similar information associated with segmentation areas. Another significant point for considering rotation is the availability of good open-source libraries that can automatically measure the pose of the head. These can be used both to guide and to test the effectiveness of the operation.

\begin{figure}[H]
    \centering
    \includegraphics[width=.9\columnwidth]{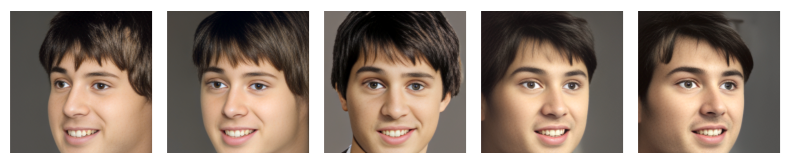}\\
    \includegraphics[width=.9\columnwidth]{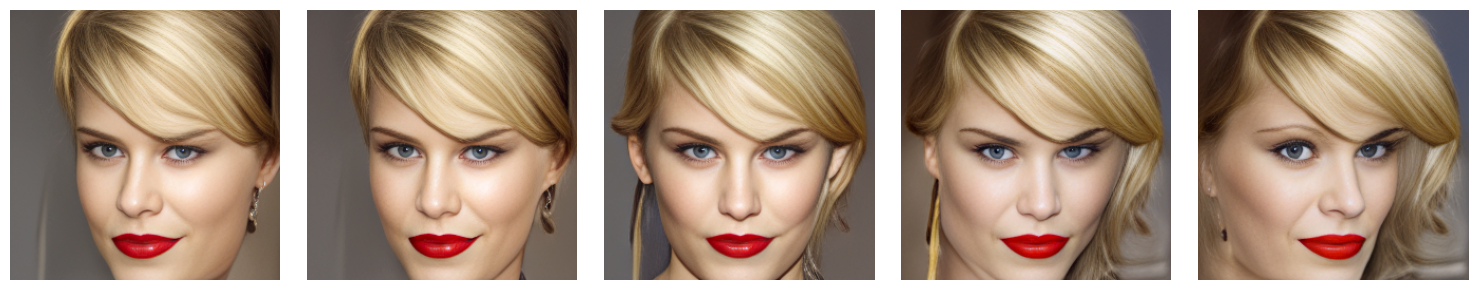}
    \includegraphics[width=.9\columnwidth]{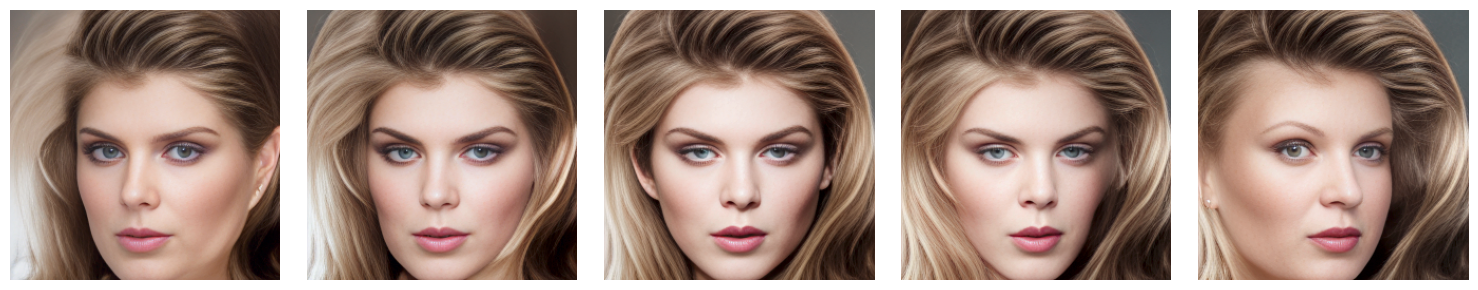}
    \caption{Rotation examples. The sources are images 114 16399, and 98018 of CelebA (central image).}
    \label{fig:example_114}
\end{figure}

By employing the DDIM embedding technique, we have been able to achieve remarkable manipulations in head orientation, spanning a large rotation angle of $\pm 30^o$ along the yaw direction. Some examples are given in Figure~\ref{fig:example_114}.
This seems to testify that compulsory trajectories can be defined in the latent space of diffusion models inspite of the intrinsically multimodal nature of the embedding function.

Our methodology involves utilizing a {\em pre-trained} generative latent model
for face generation, computing in its latent space trajectories composed of rectilinear segments simulating the rotation effect. The direction of
each segment is computed by linear regression, fitting through clouds of latent representations of dataset samples with varying yaw rotations. Each 
segment is then translated to the correct and known source location.
To obtain trajectories tailored to a specific source image, we restrict the analysis to subsets of data sharing significant attributes with it; this is
usually sufficient to ensure that the essential characteristics of the face are preserved throughout the manipulation process. 
We tested several attributes, and the most significant ones appear to be gender, expression (smiling/not smiling), age (young/old), and illumination source (left/center/right).
This last attribute is not a traditional attribute of the CelebA datasets;
we created such labeling in recent years through the collaboration
of many students, following a methodology briefly described in Section~\ref{sec:celeba}.  

The analysis and comparison of the attributes, highlighting the importance of considering the source of illumination to achieve good rotation effects, is the main contribution of our work in the specific domain of face editing.

In this article, we present our methodology and showcase some preliminary, experimental results of the manipulations performed using the DDIM embedding technique. We do not yet have a quantitative evaluation of our work, due to the
difficulty in identifying proper metrics; this is left as a subject for
further investigation. Nevertheless, our findings demonstrate the potential of DDMs in enabling intricate editing operations while maintaining the fidelity of generated samples. The insights gained from this research contribute to advancing the field of deep generative modeling and provide a valuable foundation for future developments in latent space exploration and attribute manipulation. 

The article is structured in the following way. In Section~\ref{sec:related_works} we discuss related works, and clarify 
the scope of our research aimed to understand the dynamic of head movement in the latent space of Diffusion Models. 
Section~\ref{sec:theory} briefly presents the theory of this class
of generative models; this section does not contain original material
and can be skipped by readers knowledgeable in the area.
In Section~\ref{sec:models}, we discuss the architecture of the neural 
models used for our work. Section~\ref{sec:celeba} introduces the CelebA dataset, its attributes, and our original labeling relative to the illumination source. In Section~\ref{sec:methodology}, we explain our methodology. Preprocessing operations (cropping and background removal) and postprocessing ones (super resolution and 
color correction) are discussed in Section~\ref{sec:pre_post_processing}. Numerous examples are given in Section~\ref{sec:results} and in the supplementary material. Moreover, in this section, we particularly focus on the current limitations of the technique, presenting a large number of problematic cases. Finally, concluding remarks and ideas for future research directions are given in Section~\ref{sec:conclusions}.

\section{Related Works}\label{sec:related_works}
The task of head rotation holds significant importance in computer vision, finding extensive applications in various domains like security, entertainment, and healthcare.

Before the rise of deep learning, facial rotation methods primarily revolved around applying the traits of an input face image onto a 3D face model and then rotating it to create the desired rotated version. Examples of this approach can be found in \cite{rotation_model3d_1} and \cite{rotation_model3d_2}. In \cite{rotation_model3d_3}, 
the rotation problem was tackled using a 3D transformation matrix, which mapped each point from a 2D face image to its corresponding point on a 3D face model. Although these techniques could generate rotated face images, they were constrained by distortion and blurring issues that arose during the conversion of 2D images into 3D models.

The progress of deep learning has significantly expedited facial rotation techniques, especially those leveraging generative adversarial networks (GANs). A typical
application is face frontalization, aiming to improve face recognition accuracy by synthesizing a frontal face image from a side-view facial image.

Popular techniques in this category include DR-GAN \cite{DR-GAN}, TP-GAN \cite{TP-GAN}, CAPG-GAN \cite{CAPG-GAN}, and FNM \cite{FNM}. 
DR-GAN isolates the input image's features and angle to generate a frontal image, whereas TP-GAN separately learns the overall outline features and detailed features to synthesize the frontal face. CAPG-GAN utilizes a heat map to frontalize an input face, and FNM leverages both labeled and unlabeled data to improve learning efficiency

All these methods face challenges in producing convincing results for input images taken from near-side angles or angles that are not frontal views.

Several 3D geometry-based approaches have been devised to tackle head rotation challenges by combining traditional techniques with GANs.
Relevant methods in this domain include FF-GAN \cite{FF-GAN}, UV-GAN \cite{UV-GAN}, HF-PIM \cite{HF-PIM}, and Rotate-and-Render \cite{Rotate-and-render}. These techniques leverage the strengths of both 3D modeling and GANs to achieve more realistic and accurate rotations, overcoming some limitations of purely 2D or GAN-only approaches.              

In contrast to reconstruction-based techniques, 3D geometry-based methods produce more realistic results for side-facing images. However, the need to handle detailed geometrical data, perform extensive rendering calculations, and integrate multiple complex processes makes 3D geometry-based methods more resource-intensive compared to other generative techniques.

Neural Radiance Fields (NeRF) \cite{nerf} is an advanced method for representing intricate 3D scenes by means of neural networks. NeRF models the radiance and volume density of a scene as a continuous function. This function is parameterized by a neural network that receives a 3D coordinate and a viewing direction as inputs. The scene's appearance is rendered by integrating the radiance along each camera ray. 
In \textit{FENeRF: Face Editing in Neural Radiance Fields}, the authors utilize NeRF to forecast a 3D representation of a given face with a particular rotation. This representation can then be further manipulated to edit the facial attributes. All these approaches are sensibly different from our work, since they aim to train conditional models taking into account geometric or texture information constraining the generation. In our case, we simply start from a unconstrained generative model, already containing in its latent space
the source and target image, and try to identify the path leading from the
source to the target. The purpose of this research is to investigate the structure of the latent space of generative models, to better understand the learned representation, and the properties of encodings. 

Several works have been done in 
this direction in the case of GANs,
aimed to manipulate and govern the attributes of generated faces through a latent space-based approach. These techniques enable the control over various attributes, including age, eyeglasses, gender, expression, and rotation angles of the synthesized faces. 
Different methods have been developed, including PCA analysis to extract important latent directions \cite{ganLatentSpace1}, semantic analysis to control various attributes \cite{ganLatentSpace2}, and composing a new latent vector to control multiple attributes \cite{ganLatentSpace3}. 
The most recent research mostly focused on text-guided image editing
\cite{textualInversion,GalPMBCC22}, frequently exploiting segmentation masks to drive generation \cite{MoritaZHZ23}.

All methods address rotation as a single shot operation, 
failing to provide evidence of a smooth and continuous modification of the source along a given trajectory.

Similar works have been done in the case of Variational Autoencoders (VAEs).
In this context, due to the Gaussian-like shape of the latent space induced by the Kullback-Leibler regularization, more principled approaches to the computation of trajectories can be considered, for instance considering
geodesic paths \cite{kalatzis2020variational, chadebec2022geometric,shamsolmoali2023vtae}. In the case of DDMI, the source
space is indeed Gaussian too, but this is just the source noisy space 
rapidly collapsing, after a few iterations of the denoising process, 
towards the actual manifold of data. So there is no evidence that following
a geodesic path could be beneficial, and out investigation seems
to suggest that this is not the case (see Section \ref{sec:slopes_comparison}).

In the specific case of Diffusion Models, there are several recent investigations on text-guided generation 
\cite{Prompt-to-prompt,SINE,Imagic,DiffEdit}, but we are aware of no work focused on trajectories for continuous transformations, as the ones addressed by our research.

\section{Denoising Diffusion Models}
\label{sec:theory}
This section provides a fairly self-contained theoretical 
introduction to diffusion models. It has been added for the sake of completeness, and for introducing the terminology. It does not contain original material, and it can be skipped by people with knowledge in the domain. 
We refer the readers to these excellent textbooks for additional information \cite{hands_on2024,bishop2023diffusion}.

\subsection{Diffusion and reverse diffusion}\label{sec:diffusion}
Suppose to have data distributed according to some probability distribution
$x_0 \sim q(x_0)$. We consider a forward process which
gradually adds noise to the data, producing noised samples $x_1, \dots, x_T$, for some time horizon $T > 0$. Specifically, the diffusion model $q(x_{0:T})$ is supposed to be a Markov chain with the following shape:
\begin{equation}
    q(x_t \vert x_{t-1}) = \mathcal{N}\Biggl(x_t \Bigg\vert \sqrt{\frac{\alpha_t}{\alpha_{t-1}}} x_{t-1}; \Bigl(1 - \frac{\alpha_t}{\alpha_{t-1}}\Bigl) \cdot I\Biggr)
\end{equation}
with $\{ \alpha_t \}_{t \in [0, T]}$ being a decreasing sequence in the interval $[0, 1]$. 

Considering the fact that the composition of Gaussian distributions is still Gaussian, in order to sample $x_t \sim q(x_t\vert x_0)$ we do not need to go through an iterative process. If we define $\alpha_t = 1 -\beta_t$ and $\overline{\alpha}_t = \prod_{s=0}^t\alpha_s$, then
\begin{align*}
q(x_t\vert x_0) & = \mathcal{N}(x_t;\sqrt{\overline{\alpha}_t}x_0,(1-\overline{\alpha}_t)\bf{I})\\
& = \sqrt{\overline{\alpha}_t}x_0 + \epsilon \sqrt{1 -\overline{\alpha}_t} \tag{\theequation}\label{eq:direct_flow}
\end{align*}
for $\epsilon \sim \mathcal{N}(0,\bf{I})$. In these equations, 
$1-\overline{\alpha}_t$ is the variance of the noise for an arbitrary time step $t$; $\overline{\alpha}_t$ could be equivalently used instead of $\beta_t$ to define the schedule of the noising process.

The idea behind denoising generative models is to reverse the above process, addressing the distribution $q(x_{t-1}\vert x_t)$. If we know how to sample from $q(x_{t-1}\vert x_t)$, then we can generate a sample starting from a Gaussian noise input $x_{T} \sim \mathcal{N}(0,\bf{I})$.
In general, the distribution $q(x_{t-1}\vert x_t)$ cannot be expressed in closed form, and it will be approximated using a neural network. 
In \cite{Sohl-DicksteinW15} it was observed that $q(x_{t-1}\vert x_t)$ approaches a
diagonal Gaussian distribution when $T$ is large and $\beta_t \to 0$, 
so in order to learn the distribution it suffices to train a
neural network predicting the mean $\mu_\theta$ and the diagonal covariance matrix 
$\Sigma_\theta$: 
\[p_\theta(x_{t-1}\vert x_t) = \mathcal{N}(x_{t-1}; \mu_\theta(x_t, t), \Sigma_\theta(x_t, t))\]
The whole {\em reverse} process is hence:
\begin{equation}
    p_\theta (x_{0:T}) = p_\theta(x_T) \prod_{t=1}^T p_\theta (x_{t-1} \vert x_t)
\end{equation}
where $p_\theta(x_T) = \mathcal{N}(0,\bf{I})$.

For training, we can use a variational lower bound on the negative log likelihood:
\[
\begin{array}{l}
- \log p_\theta(\mathbf{x}_0) \\
\hspace{.7cm}\leq - \log p_\theta(x_0) + D_\text{KL}(q(x_{1:T}\vert x_0) \| p_\theta(x_{1:T}\vert x_0) ) \\
\hspace{.7cm}= -\log p_\theta(x_0) + \mathbb{E}_{x_{1:T}\sim q(x_{1:T} \vert x_0)} \Big[ \log\frac{q(x_{1:T}\vert x_0)}{p_\theta(x_{0:T}) / p_\theta(x_0)} \Big] \\
\hspace{.7cm}= -\log p_\theta(x_0) + \mathbb{E}_q \Big[ \log\frac{q(x_{1:T}\vert x_0)}{p_\theta(x_{0:T})} + \log p_\theta(x_0) \Big] \\
\hspace{.7cm}= \mathbb{E}_q \Big[ \log \frac{q(x_{1:T}\vert x_0)}{p_\theta(x_{0:T})} \Big] \\
\hspace{.7cm}= \mathbb{E}_q \Big[  -\log p(x_T) - \sum_{t\ge 1} \log \frac{p_\theta(x_{t-1}\vert x_t)}{q(x_t\vert x_{t-1})} \Big] 
= \mathcal{L}(\theta)
\end{array}
\]
This can be further refined expressing $L_\theta$ as the sum of the following terms\cite{Sohl-DicksteinW15}:
\begin{equation} 
\label{eq:loss_sum}
L_\theta = L_T + L_{t-1} + \dots + L_0 
\end{equation}
where
\[
\begin{array}{l}
L_T = D_\text{KL}(q(x_T \vert x_0) \parallel p_\theta(x_T)) \\
L_t = D_\text{KL}(q(x_t \vert x_{t+1}, x_0) \| p_\theta(x_t \vert x_{t+1})) \text{ for }1 \leq t \leq T-1 \\
L_0 = - \log p_\theta(x_0 \vert x_1)
\end{array}
\]
The advantage of this formulation is that the forward process posterior $q(x_t \vert x_{t+1}, x_0) $ becomes tractable when conditioned on $x_0$ and assume a Gaussian distribution:
\begin{equation}
\label{eq:gaussian_posterior}
 q(x_{t-1} \vert x_t, x_0) = \mathcal{N}(x_{t-1} \vert \tilde{\mu}(x_t, x_0); \tilde\beta_t{\bf I})
 \end{equation}
 where
 \begin{equation}
 \label{eq:mu_tilde}
\tilde{\mu}_t (\mathbf{x}_t, \mathbf{x}_0)
= \frac{\sqrt{\alpha_t}(1 - \bar{\alpha}_{t-1})}{1 - \bar{\alpha}_t} \mathbf{x}_t + \frac{\sqrt{\bar{\alpha}_{t-1}}\beta_t}{1 - \bar{\alpha}_t} \mathbf{x}_0
\end{equation}
and 
\begin{equation}
\tilde{\beta}_t =  \frac{1 - \bar{\alpha}_{t-1}}{1 - \bar{\alpha}_t} \cdot \beta_t.
\end{equation}
In consequence of this, the KL divergences in Eq.~\ref{eq:loss_sum} are comparisons between Gaussians, and they can be calculated in a Rao-Blackwellized fashion with closed form expressions.

After a few manipulations, we get:
\begin{equation}\label{eq:ELBO_formulation}
    \mathcal{L}(\theta) = \sum_{t=1}^T \gamma_t \E_{q(x_t \vert x_0)} \Bigl[ \| \mu_\theta(x_t, \alpha_t) - \tilde{\mu}(x_t, x_0) \|_2^2 \Big]
\end{equation}
that is just a weighted mean squared error between the image produced from $p_\theta(x_t \vert x_0)$ and the true image given by the reverse diffusion process $q(x_{t-1} \vert x_t, x_0)$ for each time $t$. 

In \cite{DDIM},
they just use a slightly different approach based on predicting the noise
$\epsilon_\theta(x_t,t)$ in a given image $x_t$, instead of denoising it.

Recall that the purpose of the training network is to approximate 
the conditioned probability distributions of the reverse diffusion process:
\[p_\theta(x_{t-1}\vert x_t) = \mathcal{N}(x_{t-1}; \mu_\theta(x_t, t), \Sigma_\theta(x_t, t))\]
Our goal is to train the network to predict $\tilde{\mu}$ of equation~\ref{eq:mu_tilde}. Since, $x_0 = \frac{1}{\sqrt{\bar{\alpha}_t}}(x_t - \sqrt{1 - \bar{\alpha}_t}\epsilon_t)$, we have 
\begin{equation}
\label{q:mu_theta}
\mu_\theta(x_t, t) = \frac{1}{\sqrt{\alpha_t}} \Big(x_t - \frac{1 - \alpha_t}{\sqrt{1 - \bar{\alpha}_t}} \epsilon_\theta(x_t, t) \Big)
\end{equation}
and
\begin{equation}
\label{eq:x_t_misnus1}
x_{t-1} \sim \mathcal{N}(\mathbf{x}_{t-1}; \frac{1}{\sqrt{\alpha_t}} \Big( \mathbf{x}_t - \frac{1 - \alpha_t}{\sqrt{1 - \bar{\alpha}_t}} \epsilon_\theta(\mathbf{x}_t, t) \Big), \Sigma_\theta(\mathbf{x}_t, t))
\end{equation}
The network can be simply trained to minimize the quadratic distance between the actual and the predicted error. Ignoring weighting terms, that seems to be irrelevant if not harmful in practice, the loss is just:
\begin{equation}
L_t^\text{simple} = \mathbb{E}_{t \sim [1, T], x_0, \epsilon_t} \Big[\| \epsilon_t - \epsilon_\theta(x_t, t) \|^2 \Big]
\end{equation}
$L_t^\text{simple}$ does not give any learning signal for $\Sigma_\theta(x_t, t)$. In \cite{DDIM}, the authors preferred to fix it to a constant, testing both $\beta_t{\bf I}$ and $\tilde{\beta}_t{\bf I}$, with no sensible
difference between the two alternatives.

\subsection{Pseudocode}
With the above setting, the algorithms for training and sampling are very
simple. The network $\epsilon_\theta(x_t, t)$ takes in input a noisy image
$x_t$ and time step $t$, and it is supposed to return the noise contained in the image. Suppose to have a given noise scheduling $(\alpha_T, \dots, \alpha_1)$. We can train the network in a supervised way, sampling a
true image $x_0$, creating a noisy version of it  $x_t = \sqrt{\alpha_t} x_0 + \sqrt{1\!-\! \alpha_t} \epsilon$ where $\epsilon \sim \mathcal{N}(0;I)$, and instructing the network to guess $\epsilon$.

Note that we only have a single network, parametric in the time step $t$ (or, since it is equivalent, parametric in $\alpha_t$).

\begin{algorithm}[H]
\caption{Training}
\label{algorithm1}
\begin{algorithmic}[1]
    \State Fix a noise schedule $\{\alpha_t\}_{t=1}^T$
    \Repeat
        \State Sample $x_0 \sim P_{\text{data}}$
        \State Sample $t \sim \text{Uniform}(\{1, \dots, T\})$
        \State Sample $\epsilon \sim \mathcal{N}(0, I)$
        \State Compute $x_t = \sqrt{\alpha_t}\, x_0 + \sqrt{1 - \alpha_t}\, \epsilon$
        \State Take gradient descent step on $\left\| \epsilon - \epsilon_\theta(x_t, t) \right\|^2$
    \Until{converged}
\end{algorithmic}
\end{algorithm}

Generative sampling is performed through an iterative loop: we start from a purely noisy image $x_T$ and progressively remove
noise by means of the denoising network. The denoised version of the 
image at time step $t$ is obtained from Eq.~\ref{eq:x_t_misnus1}.

\begin{algorithm}[H]
\caption{Sampling}
\label{algorithm2}
\begin{algorithmic}[1]
    \State Sample $x_T \sim \mathcal{N}(0, I)$
    \For{$t = T, \dots, 1$}
        \State Sample $z \sim \mathcal{N}(0, I)$ \textbf{if} $t > 1$ \textbf{else} $z = 0$
        \State Compute 
        \[
        x_{t-1} = \frac{1}{\sqrt{\alpha_t}} \left( x_t - \frac{1 - \alpha_t}{\sqrt{1 - \bar{\alpha}_t}} \, \epsilon_\theta(x_t, t) \right) + \sigma_t z
        \]
    \EndFor
\end{algorithmic}
\end{algorithm}


Several improvements can be made to this technique.

A first important point concerns the noise scheduling $\{ \alpha_t \}_{t=1}^T$.
In \cite{DDPM}, the authors used linear or quadratic schedules.
This typically results in a vary steep decrease during the initial time steps, that could be problematic for generation. 
In order to address this issue, alternative scheduling functions that incorporate a more gradual decrease, such as the 'cosine' or 'continuous cosine' schedule, have been proposed in the literature \cite{nichol2021improved}.
The precise choice of the scheduling function does not seem to matter, provided it shows a nearly linear behaviour in the middle of the generative process and smoother changes around the beginning and the end of the scheduling.

Another major issue regards the speedup of the sampling process.
that in the original approach was up to one or a few thousand steps.
Since the generative model approximates the reverse of the inference process, in order to reduce the number of iterations required by the generative model, it could be worth rethinking the inference process. This investigation motivated the definition of Denoising Deterministic Implicit Models, explained in the following 
section.

\subsection{Denoising Deterministic Implicit Models}
Denoising Deterministic Implicit Models (DDIMs) \cite{DDIM} are a variation of the previous approach exploiting a non-Markovian noising process
having the same forward marginals as DDPM, but allowing a better tuning of the variance of the reverse noise. 

We start by making the definition of $q(x_{t-1} \vert x_t, \mathbf{x}_0)$ parametric with respect to a
desired standard deviation $\sigma$:
\begin{align}
x_{t-1} 
&= \sqrt{\bar{\alpha}_{t-1}}x_0 +  \sqrt{1 - \bar{\alpha}_{t-1}}\epsilon_{t-1} \\
&= \sqrt{\bar{\alpha}_{t-1}}x_0 + \sqrt{1 - \bar{\alpha}_{t-1} - \sigma_t^2} \epsilon_t + \sigma_t\epsilon \\
&= \sqrt{\bar{\alpha}_{t-1}}x_0 + \sqrt{1 - \bar{\alpha}_{t-1} - \sigma_t^2} \frac{x_t - \sqrt{\bar{\alpha}_t}x_0}{\sqrt{1 - \bar{\alpha}_t}} + \sigma\epsilon
\end{align}
So, 
\begin{align}
\label{eq:x_tminus1_sampling}
q_\sigma(x_{t-1} \vert x_t, x_0)
&= \mathcal{N}(x_{t-1}; \mu_{\sigma_t}(x_0, \alpha_{t-1}), \sigma_t^2 \mathbf{I})
\end{align}
with 
\begin{align}
    \mu_{\sigma_t}(x_0, \alpha_{t-1}) = \sqrt{\bar{\alpha}_{t-1}}x_0 + \sqrt{1 - \bar{\alpha}_{t-1} - \sigma_t^2} \frac{x_t - \sqrt{\bar{\alpha}_t}x_0}{\sqrt{1 - \bar{\alpha}_t}}
\end{align}
According to this approach, the forward process is no longer Markovian, but it depends both on the starting point $x_0$ and on $x_{t-1}$. However, it can be easily proved that the marginal distribution $q_\sigma(x_t\vert x_0) = \mathcal{N}(x_t \vert \sqrt{\overline{\alpha}_t} x_0; (1-\bar{\alpha}_t) \cdot I)$ recovers the same marginals as in DDPM. As a result, $x_t$ can be diffused from $x_0$ and $\alpha_t$ by generating a realization of normally distributed noise $\epsilon_t \sim \mathcal{N}(\epsilon_t \vert 0; I)$.  

We can set $\sigma_t^2 = \eta \cdot \tilde{\beta}_t $ where $\eta$ is a control parameter that can be used to tune sampling stochasticity. In the special case 
when $\eta=0$, the sampling process becomes completely deterministic.

The sampling procedure in case of DDIM is slightly different from the case of 
DDPM. In order to sample $x_{t-1}$ according to Eq.~\ref{eq:x_tminus1_sampling}, we need $x_0$, that
is obviously unknown at generation time. Since, however, at each step we are guessing the amount of noise $\epsilon(x_t,t)$ in $x_t$, we can generate a denoised observation $\tilde{x}_0$, which is a prediction of $x_0$ given $x_t$:
\[\tilde{x}_0 = (x_t - \sqrt{1-\bar{\alpha}_t}\epsilon_\theta(x_t,t)/\sqrt{\bar{\alpha}_t)}\]
The full generative algorithm is summarized in the following pseudocode:

 \begin{algorithm}[H]
\caption{Sampling}
\label{algorithm3}
\begin{algorithmic}[1]
    \State Sample $x_T \sim \mathcal{N}(0, I)$
    \For{$t = T, \dots, 1$}
        \State Compute $\epsilon = \epsilon_\theta(x_t, \bar{\alpha}_t)$
        \State Compute 
        \[
        \tilde{x}_0 = \frac{1}{\sqrt{\bar{\alpha}_t}} \left( x_t - \frac{1 - \bar{\alpha}_t}{\sqrt{1 - \bar{\alpha}_t}} \, \epsilon \right)
        \]
        \State Compute 
        \[
        x_{t-1} = \sqrt{\bar{\alpha}_{t-1}}\, \tilde{x}_0 + \sqrt{1 - \bar{\alpha}_{t-1}}\, \epsilon
        \]
    \EndFor
\end{algorithmic}
\end{algorithm}

An interesting aspect of DDIM, frequently exploited in the literature, is that 
due to its deterministic nature, it also defines an
{\em implicit} latent space. This opens the way to very interesting operations comprising latent space interpolation, or the exploration of interesting trajectories for editing purposes. The latent space can be obtained by integrating an ODE in the forward direction and then reverse the process to get the latent encodings that produce a given real image \cite{Diff_vs_GAN}. In \cite{asperti2022embedding}, it was shown that a deep neural network can also be trained to perform this embedding operation, sensibly reducing
its cost. We shall provide details on the embedding network in Section~\ref{sec:embedding_network}.

\section{Model architecture}
\label{sec:models}
As made clear in the previous Section, the main component of a diffusion model is a denoising network, that takes in input a noise variance $\bar{\alpha}_t$, an image $x_t$ corrupted with a corresponding amount of noise, and tries to guess the actual noise
$\epsilon_\theta(x_t,\bar{\alpha}_t)$ present in the image. Starting from an image $x_0$ of the data distribution, we can generate a random noise $\epsilon \sim \mathcal{N}(0;I) $, and produce a corrupted version $x_t =  \sqrt{\bar{\alpha}_t} x_0 + \sqrt{1\!-\! \bar{\alpha}_t} \epsilon$. Then, the network is trained to minimize the distance between 
the actual noise $\epsilon$, and the predicted one $\epsilon_\theta(x_t,\bar{\alpha}_t)$:
\begin{equation}
Loss = \| \epsilon - \epsilon_\theta(x_t, \bar{\alpha}_t) \|^2 
\end{equation}

The architecture of this network is traditionally based
on the U-net \cite{U-net}, a well known convolutional neural network introduced, in origin, for image segmentation tasks. 

\begin{figure}[h]
    \centering
    \includegraphics[width=.9\columnwidth]{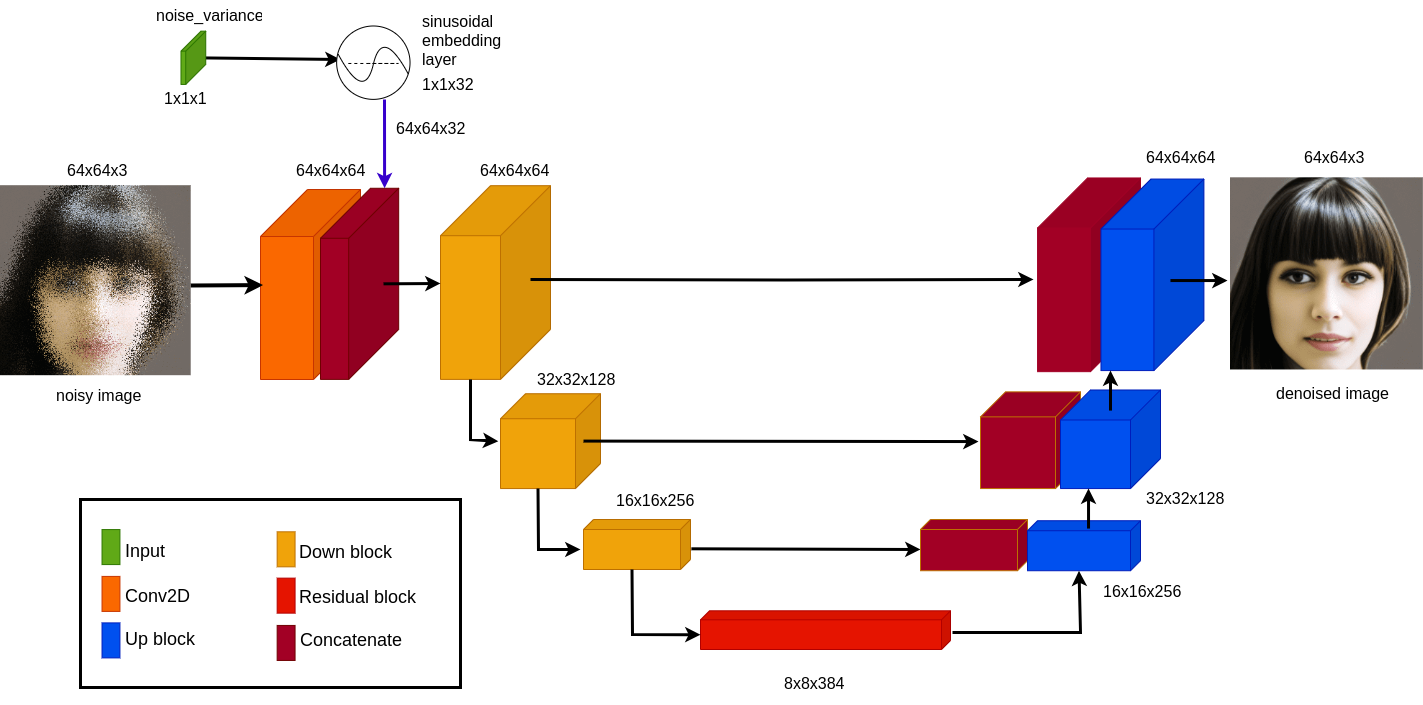}
  \caption{The U-net architecture of our denoising model. \label{fig:U-net}} 
\end{figure}

The U-net (see Figure~\ref{fig:U-net}) features an encoder-decoder structure, incorporating skip connections between layers of the encoder and decoder with corresponding spatial dimensions. We work on images with an initial resolution $64\times 64$  \cite{asperti2022embedding}.

\begin{figure}[h]
    \centering
    \includegraphics[width=.9\columnwidth]{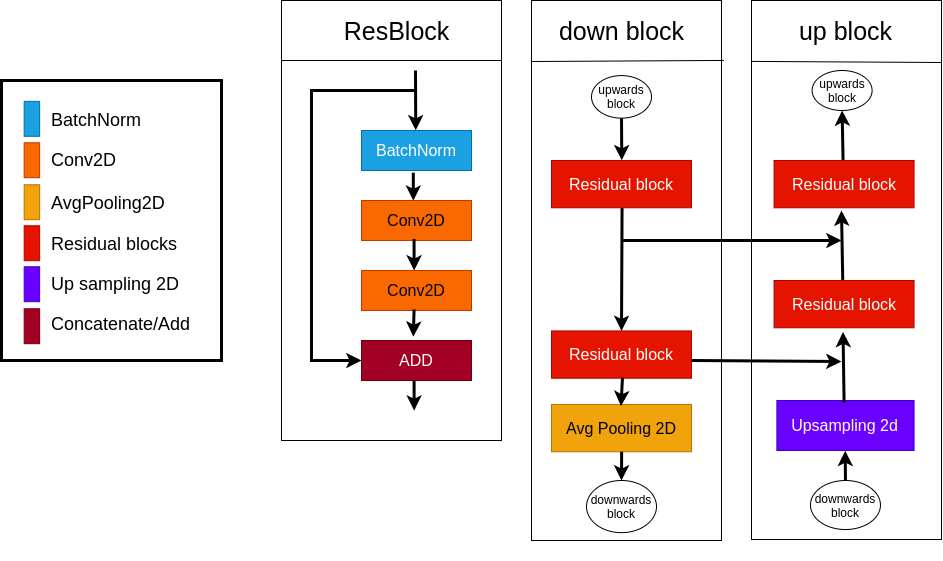}
  \caption{Main architectural modules, including the
residual block, down block and up block.\label{fig:U-net-modules}} 
\end{figure}

The input relative to the noise variance $\alpha_t$ is typically embedded using
sinusoidal position embeddings. Then, this information is vectorized and concatenated to the initial one. The detailed structures of the various modules
is given in Figure~\ref{fig:U-net-modules}.

\subsection{The embedding Network}
\label{sec:embedding_network}
In DDIM, the sampling process is deterministic given the initial noise $x_T$, so
it is natural to try to address the reverse problem, computing $x_T$ from $x_0$.
The operation is not obvious, however, since the problem is clearly underconditioned, and we may have many different points in the latent space generating the same
output. 

The embedding problem for Denoising Models can be addressed in several different 
ways: by gradient ascent, integrating the ordinary differential equations (ODE) 
defining the forward direction, and then running the process in reverse to get the latent representations, or directly trying to train a network to approximate the 
embedding task \cite{asperti2022embedding}. The advantage of the latter approach is
that, once training is completed, the computation of the latent encoding is 
particularly fast.

The embedding network $\mathit{Emb(x)}$ takes in input an image and try to compute its embedding. It is trained in a completely supervised way: given
some noise $x_T$, we generate a sample $x_0$ and train the network to 
synthesize $x_T$ from $x_0$, using as loss the distance between $x_T$ and 
$\mathit{Emb(x_0)}$. Modern neural computation environments enable us to backpropagate gradients through the iterative loop of the reverse diffusion process effortlessly.

Many different models of Embedding Network were compared in \cite{asperti2022embedding}; the best results have been obtained by a U-Net, essentially identical to the denoising network. The big difference with respect to generation, is that we compute the latent representation with a single pass through the network. 
Reconstruction has an MSE around 0.0012 in case of CelebA, that is definetely good.


\section{CelebFaces Attributes Dataset}
\label{sec:celeba}
The CelebFaces Attributes dataset (CelebA) \cite{celeba} consists of over 200,000 celebrity images, with a rich set of annotations. In the past, 
it has been widely used in the fields of computer vision and machine learning, for tasks like attribute prediction, facial recognition and generation. Each image is equipped with 40 binary attributes, covering characteristics like gender, age, hair color and so on. Additionally, bounding box annotations for the faces are provided.

CelebA exhibits a wide diversity in image quality, resolution, and sources, capturing a broad spectrum of ethnicities, age groups, and genders. To facilitate deep learning models' development, an aligned version of the dataset is offered, where faces are centered in a common coordinate system, ensuring consistent size and orientation.

CelebAMask-HQ \cite{CelebAMask-HQ} is a recent derivative of CelebA. 
It contains 30,000 high-resolution images, manually annotated with segmentation masks
relative to 19 different facial components and accessories. This dataset serves as a valuable resource for training and evaluating face parsing, recognition, generation and editing.


\subsection{Analysis of the dataset}\label{sec:Analysis of Annotations}
Since a generative model is designed to model the distribution of data, it is always a good practice to begin the study of a model with an analysis of the underlying dataset. Any bias in the data, such as an imbalance between different classes, will likely be reflected by the model, potentially leading to unexpected generative behaviors. For example, in the CelebA dataset, there is a noticeable gender imbalance, with a majority of female images over male images. Consequently, when editing faces, this bias can cause the model to more frequently transform male faces into female ones.

We compared and selected the relevant attributes for our investigation using the methodology described in Section~\ref{sec:slopes_comparison}. This approach aimed to assess the impact of each attribute on the direction of the trajectory. Ultimately, we focused on a subset of attributes that includes gender, age, smiling, and ``mouth slightly open."

\begin{figure}[H]
\centering
\includegraphics[width=.7\columnwidth]{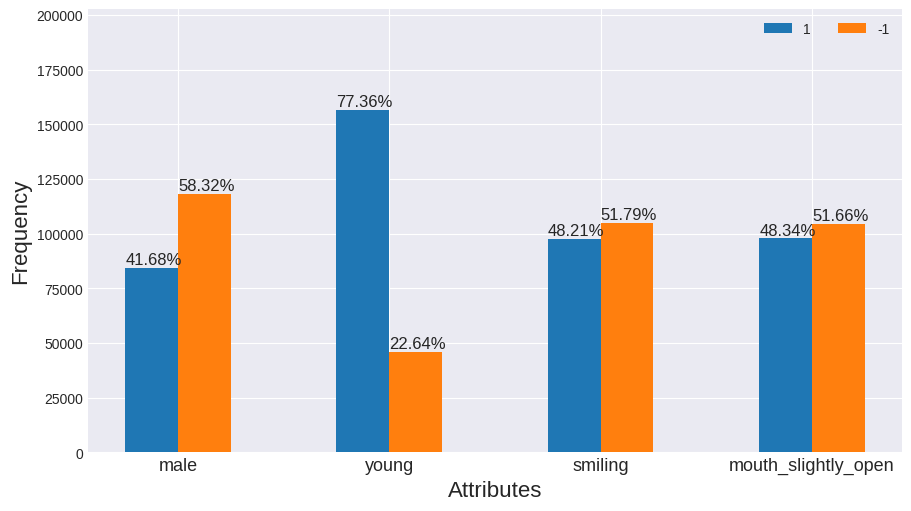} \caption{\label{fig:celeba_attributes_analysis}
Distribution of CelebA attributes. In CelebA, each attribute is annotated with either -1 or 1. For example, for gender, ``male = 1" stands for male, and ``male = -1" stands for female.}   
\end{figure}

In this section, we investigate the distribution of these attributes in CelebA; results are summarized in Figure~\ref{fig:celeba_attributes_analysis}.

With respect to some of the attributes, the dataset is highly unbalanced: approximately 58\% of the images in the dataset depict females, and around 77\% features young people. This imbalance could adversely affect the generative process, particularly when dealing with male and older poeple as compared to female and younger ones. 

In the following sections, we provide further information relative to face orientation and illumination source, not covered among CelebA attributes. 

\subsection{Face Orientation} \label{Analyzing Face Orientation}
\begin{figure}[H]
\begin{tabular}{cc}
\includegraphics[width=.3\columnwidth]{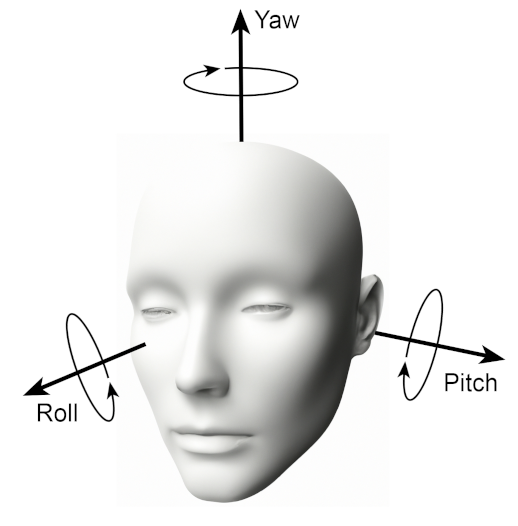} &
\includegraphics[width=.6\columnwidth]{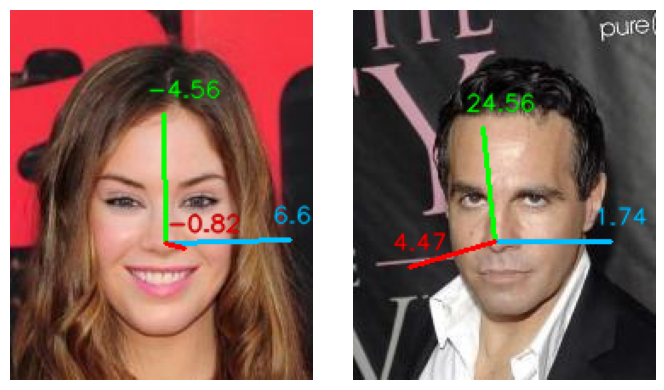}\\
(a) & (b)
\end{tabular}
\caption{\label{fig:face_orientations}
(a) Yaw, Pitch and Roll angles in HPE. (b)
Examples of head pose estimation for CelebA images: yaw is in green, pitch in blue, and roll in red.}   
\end{figure}

To guide the DDIM generation process in producing faces with different orientations, information about head orientation from the CelebA dataset is needed. This information is not included in the standard annotations of the CelebA-aligned dataset.                   
However, Head Pose Estimation is a well investigated topic \cite{HPEasperti}, and a large number of libraries are available for this purpose. In its usual formulation, the
task consists in expressing a person's head orientation in a three-dimensional space by calculating three rotation angles called yaw, pitch, and roll.  Specifically, yaw denotes the rotation around the vertical axis, pitch is the rotation around the horizontal axis, and roll is the rotation around an axis perpendicular to the other two (see Figure~\ref{fig:face_orientations})(a). 

For the straightforward task of recognizing the orientation of nearly frontal faces, as it is the case with the majority of faces in CelebA, there are open-source libraries that perform excellently. 
Specifically, we used the \texttt{cv2.solvePnP()} function from the OpenCV library \cite{opencv_library}, employing the same technique described in \cite{reification} and readily accessible from the public repository. In the same repositories, we also provide direct access to pre-computed angles for all images in the CelebA dataset.

An example of the kind of annotations which can be obtained is
given in Figure~\ref{fig:face_orientations}.

\begin{figure}[H]
\centering
\includegraphics[width=\columnwidth]{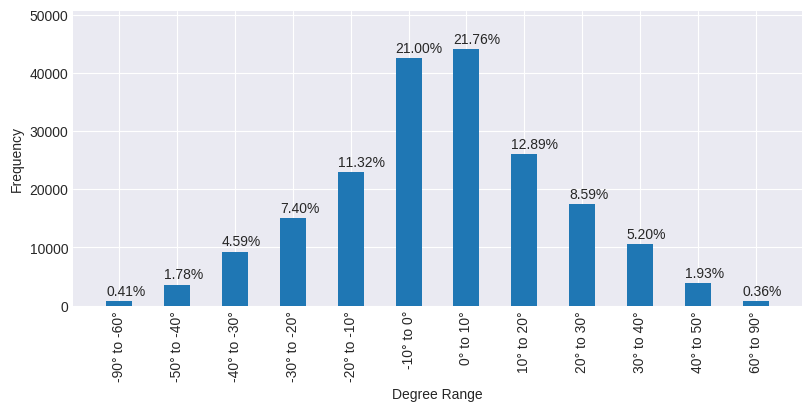} 
\caption{\label{fig:celeba_yaw_analysis}
Yaw distribution on CelebA dataset }   
\end{figure}

More interestingly, we can examine the distribution of CelebA images concerning orientations, particularly yaw, as it represents the most significant rotation in the dataset: see Figure~\ref{fig:celeba_yaw_analysis}. CelebA is an aligned dataset: as expected, over $40\%$ of the images have yaw within the $[-10, +10]$ degree range. Moreover, only $4.48\%$ of the images have yaw outside the $[-40, +40]$ degree range. The limited number of images with high yaw values restricts the generative power of the model. Typically, rotations need to be confined within a region of data with statistical significance, such as yaw in the $[-30, +30]$ degree range.


\subsection{Light Direction Analysis}
When rotating a face, it is crucial to preserve the right shadows produced by the lighting conditions. Unfortunately, no attributes are available relative to the 
source of illumination in the case of CelebA, and up to our knowledge there is no open source software able to correctly identify lighting directions in an automatic way.

An important byproduct of our work is the provision of labels for CelebA, categorizing images into three major groups based on their main source of illumination: left, center, and right. The labeling process was carried out in a semi-supervised way during the last few years with the collaboration of many students. The basic procedure involved manually annotating a large portion of the data, developing and training classification models, cross-validating the data using different models, manually revising the classifications, and repeating the process until no further critical issues emerged.


Nevertheless, this labeling has proven to be a valuable tool for our methodology, and we hope it can serve as a significant asset for further investigations into face processing tasks. The labeling can be freely accessed through the code in the github repository. We also provide pre-computed yaw, pitch and poll angles for each CelebA image.

\begin{figure}[ht]
\centering
\includegraphics[width=\columnwidth]{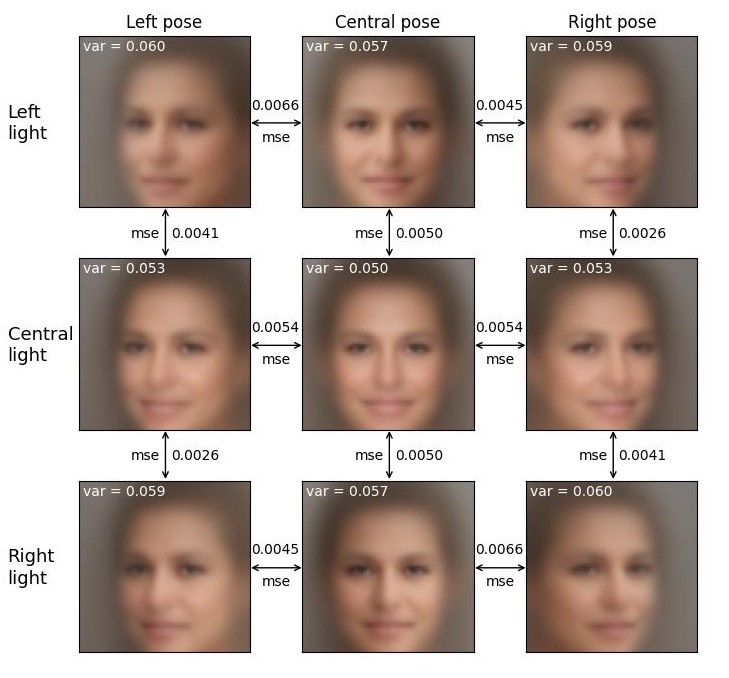}
\caption{\label{fig:mean_faces}Illumination-Pose centroids. The different figures visualize the mean faces
relative to different light sources and poses, considering three major orientation classes. We also report the variance in each class
(mean of variances of the pixels) and the mean square error (MSE) between class centers.}
\end{figure}

In Figure~\ref{fig:mean_faces}, we summarize the outcome of our labeling and the complex interplay between illumination and orientation by showing the mean faces corresponding to different light sources and poses.

In the picture, we also show the variance in each class
(mean of variances of the pixels) and the squared Euclidean distance (MSE) between class centroids.
We may observe the following points: (1) the different 
provenance of the light is still clearly recognizable in the mean faces, implicitly testifying the quality of our labeling;
(2) from the point of view of the position and shape of shadows over the phase, illumination and pose are strictly interconnected; (3) the variance of each class is an order of magnitude larger than the distance between their centers, hinting to the complexity of the classification 
problem.

The second point is particularly important for our work.
Investigations into the most relevant variables in the latent representation of images, including faces, have revealed that much of the information is conveyed by variables that explain macroattributes of the source image, such as colors and intensities of large regions (e.g., light/dark backgrounds, light/dark hair) \cite{comparingNCAA}. The intensity and positions of dark/light regions on a face are strongly influenced by the source of illumination. Therefore, it is natural to expect that this information significantly impacts the latent encoding, and indeed, as testified by this work, it does.



\section{Methodology}
\label{sec:methodology}
The problem consists of finding trajectories in latent spaces corresponding to left/right rotations of the head.

Our starting point is a large dataset of head images enriched with information related to the rotation of the head and additional attributes such as lighting source, gender, age, and expression (smiling/not smiling). The selection of these attributes is discussed in Section \ref{sec:slopes_comparison}. Images are preprocessed to remove the background as described in Section \ref{sec:pre_post_processing}.

We also assume to have a pre-trained generative model for the above dataset, along with an embedding tool capable of mapping an arbitrary sample of the dataset to its internal representation in the latent space of the generative model.

The other input is the image of the head to be rotated, let's call it $X$. Let $\Theta_X$ be its current rotation and let $Z_X$ be its latent representation. This image
can be one of the images in the dataset, or a completely new one. In the latter case, 
its current rotation, and its other attributes must be pre-computed and passed to the 
rotation model.

The methodology consists in the following steps:

\begin{description}
\item[filtering] We restrict the investigation to a subset of the dataset sharing with $X$
the selected attributes. So, if X is a young, smiling blond man with a frontal illumination,
we shall restrict the analysis to images sharing the same attributes.
\item[clustering] Starting from $\Theta_X$ we create clusters of images with rotation 
around $\Theta_X + \Delta$ for increasing values of $\Delta$ encompassing an overall rotation of $\pm 30^\circ$.
\item[embedding and centroids] we embed the clusters in the latent space and compute
their centroids. Each centroid conceptually corresponds to the latent representation of a "generic" person with the given attributes and rotations.
\item[rotation trajectories] we define rotation trajectories by fitting linear lines through the centroids using linear regression. We experimented with different spline segmentations, but ultimately obtained the best results by splitting the problem into two directions: one for right rotation and another for left rotation.
\item[re-sourcing] The final step involves applying the trajectory vector corresponding to the rotation starting from $Z_X$. We sample points along this trajectory,
and generate the corresponding images in the visible space.
\end{description}
In order to improve the quality of the final image we post-process it for super-resolution and color correction. 



\begin{figure}[ht]
    \centering
    \begin{tabular}{cc}
          \includegraphics[width=.48\columnwidth]{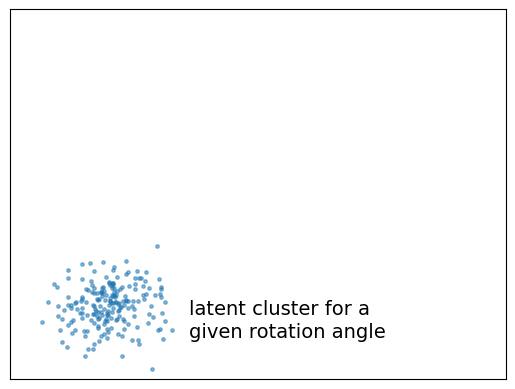} & \includegraphics[width=.48\columnwidth]{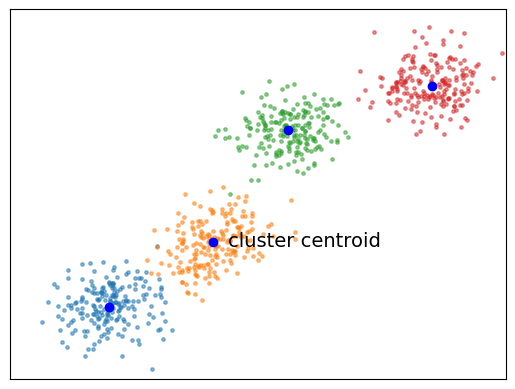} \\
          (a) clustering and embedding & (b) computing centroids\\
          \includegraphics[width=.48\columnwidth]{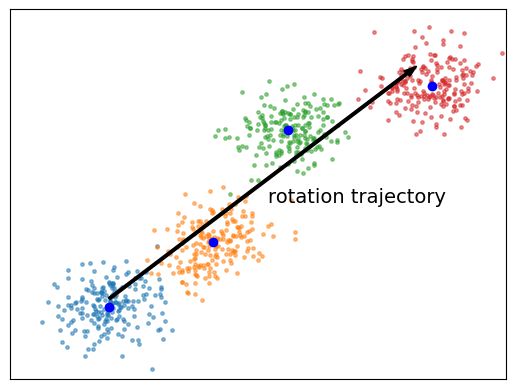} &
          \includegraphics[width=.48\columnwidth]{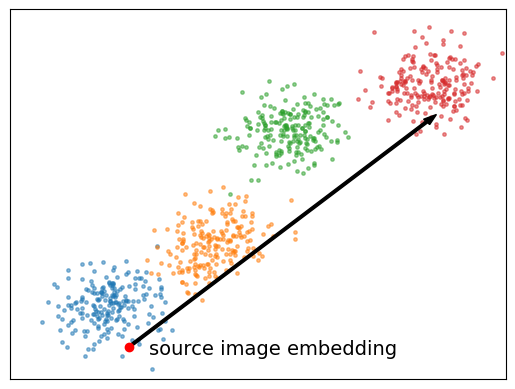} \\
          (c) fitting trajectories & (d) re-sourcing to the input 
    \end{tabular}
    \caption{Overall methodology. All pictures refer to the latent space of the generative model, schematically represented with two dimensions. We also suppose to have pre-filtered the images along the relevant attributes.
    (a) we use the embedder to computer clusters of latent points corresponding to specific rotation angles; (b) we compute the centroids of the clusters; (c) we fit a line through the centroids to compute a rotation trajectory; (d) we move along this direction starting from the specific embedding of the source images we want to rotate.}
    \label{fig:methodology}
\end{figure}

For a fixed rotation movement (left or right), the approach is schematically described in Figure~\ref{fig:methodology}. 
Our attempts to split the rotation in a larger number of linear steps
have been so far hindered by the progressive loss of the key facial characteristics of the source image.

The clustering phase is not an essential part of the algorithm, since we could 
directly apply regression on the cloud of embedded points. We compute centroids 
mostly for debugging purposes, to visualize and compare "generic" faces, for 
a given set of attributes and rotations.


The number of angles relative to centroids and their intervals can be easily customized by the user. Using a step size that is too small typically reduces the number of images retrieved from the dataset that match that specific orientation, thereby diminishing the statistical significance of the cluster. We usually work with a step size of $10^\circ$.

In the nest sections we provide additional details on some of the main steps of
our methodology.

\subsection{Filtering CelebA images}\label{Filtering_celeba}
\begin{figure}[H]
\centering
\includegraphics[width=\columnwidth]{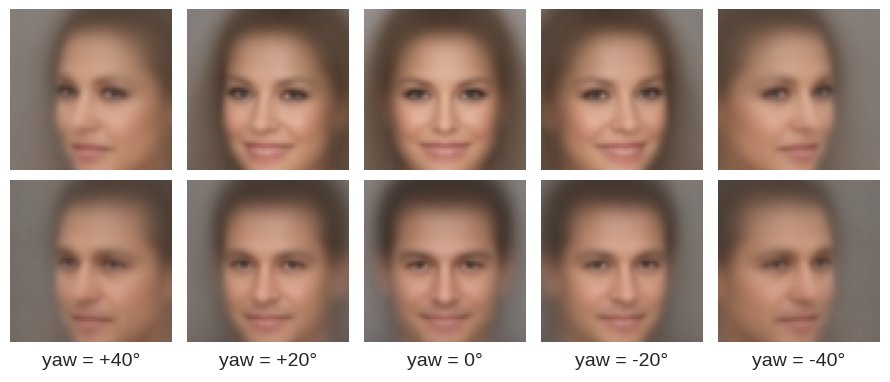}                            \caption{\label{fig:filtering_different_yaw} T
Mean images for specified yaw angles for females (top row) and males 
(bottom row).}   \end{figure}

In our first attempts, we selected images from the dataset just using the rotation.
However, it looks important to select images having at least a rough similarity
with the source image we want to act on. To this aim, we use a basic set of attributes comprising lighting source, gender, age and expression (smiling/not smiling). In Figure~\ref{fig:filtering_different_yaw} 
we show the different mean value of CelebA data relative to 
different configurations of some of these attributes.

In order to obtain a sufficiently representative number of candidates, we typically enlarge the dataset by a flipping operation, consistently inverting also the relevant attributes (yaw and lighting source). For example, if we are looking for images with a yaw of +30° a light direction of 'RIGHT', we can also take into account images with a yaw of -30° and a light direction of 'LEFT', provided we flip them. 
We aim to retrieve sets composed by at least 1K images, starting from a relatively narrow tolerance interval $[\Theta-\Delta,\Theta+\Delta]$ around the desired angle $\Theta$ and possibly enlarging $\Delta$ if required.

\begin{figure}[H]
    \centering
\begin{tabular}{lc}
& \includegraphics[width=.9\columnwidth]{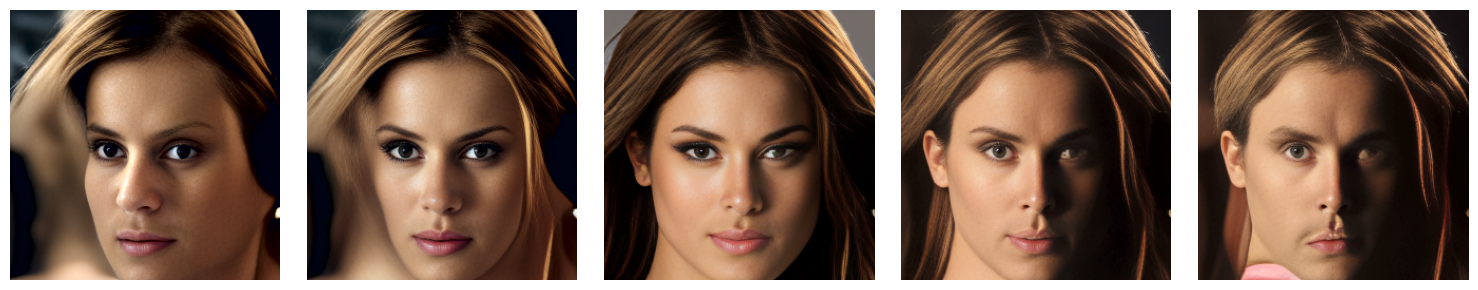}\\
(a)&\includegraphics[width=.9\columnwidth]{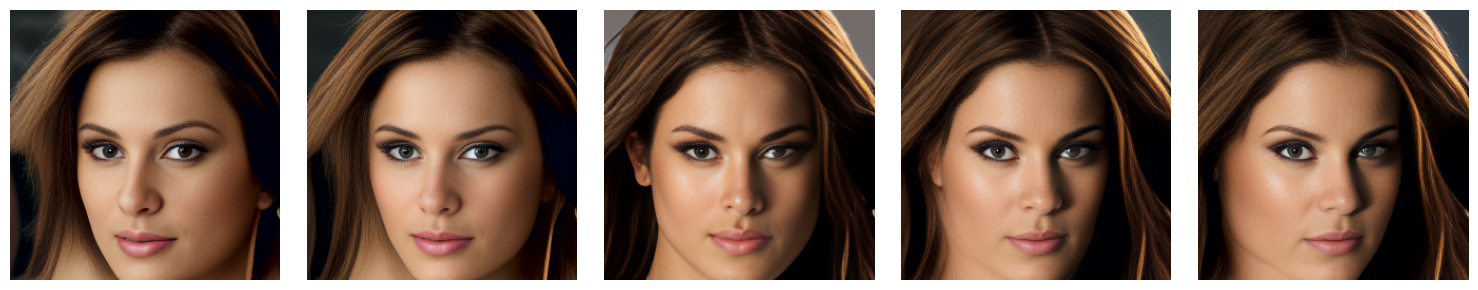}\\
\\
&\includegraphics[width=.9\columnwidth]{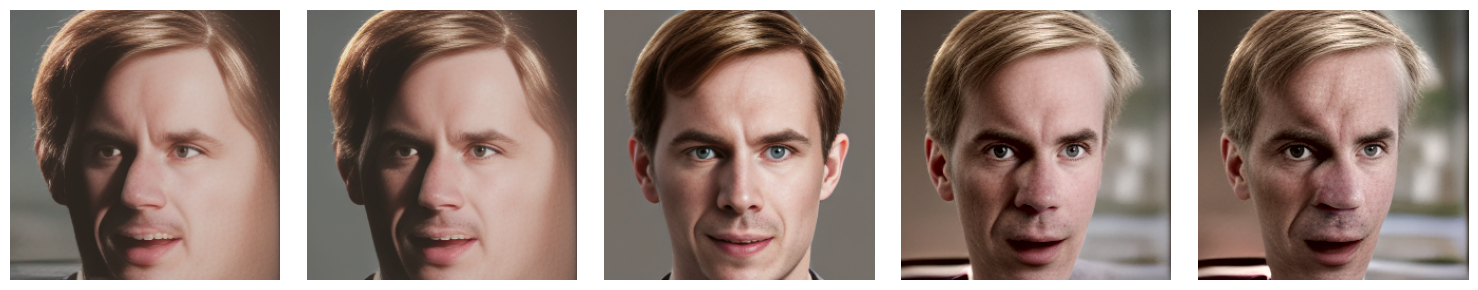}\\
(b)&\includegraphics[width=.9\columnwidth]{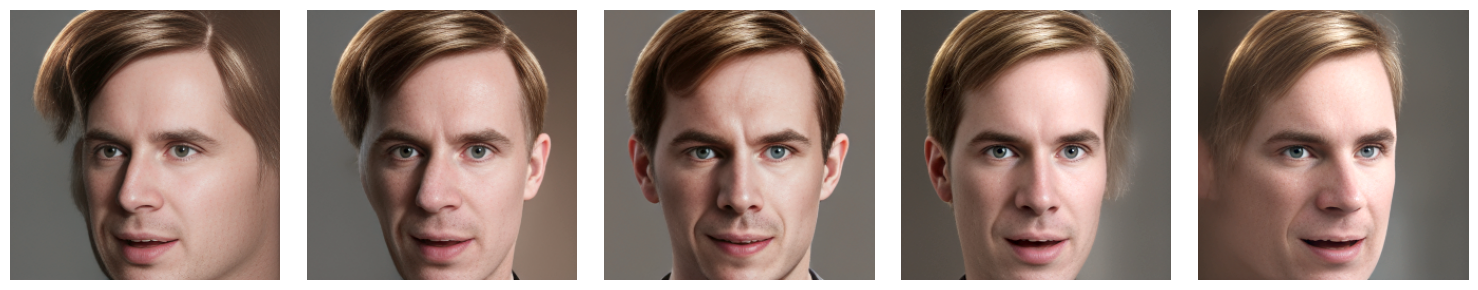}\\
\\
&\includegraphics[width=.9\columnwidth]{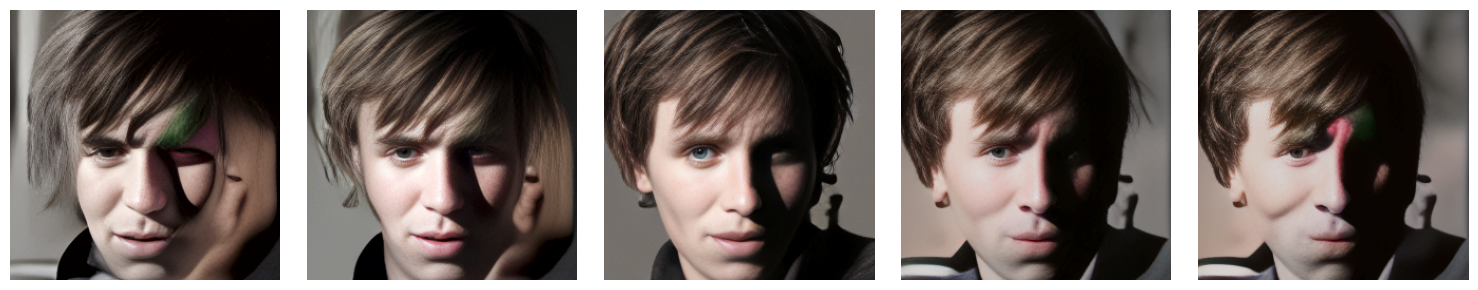}\\
(c)&\includegraphics[width=.9\columnwidth]{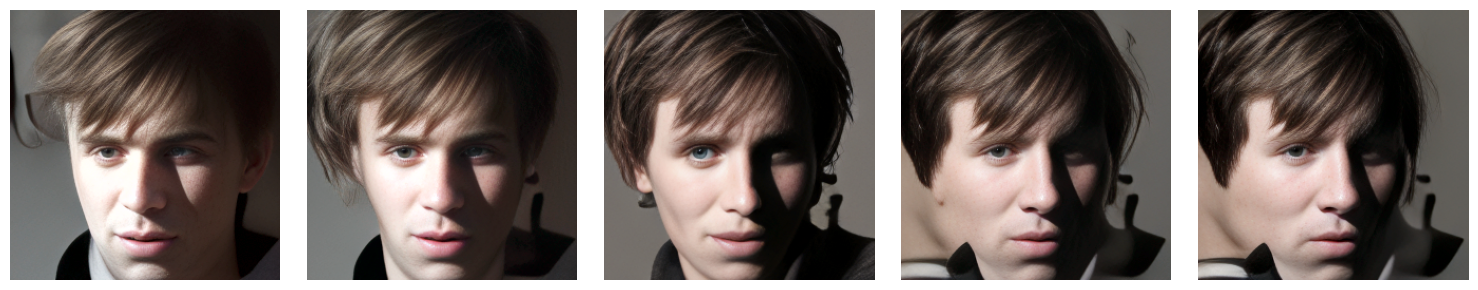}\\
\\
&\includegraphics[width=.9\columnwidth]{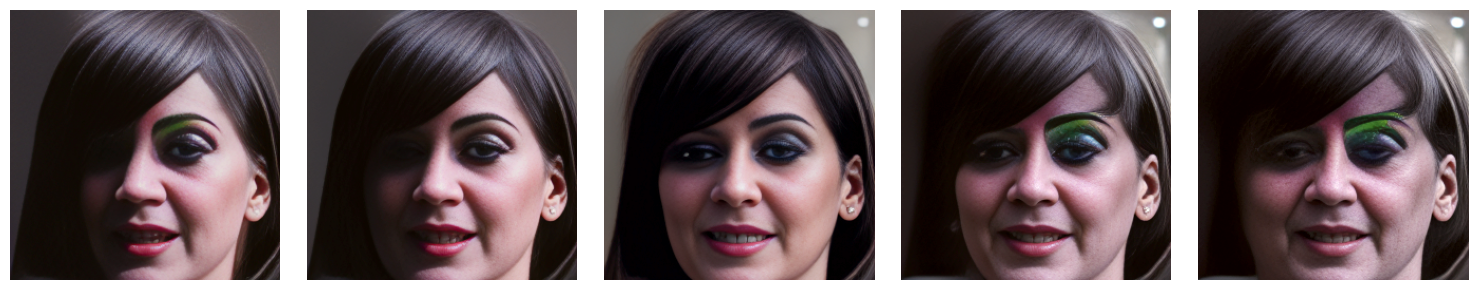}\\
(d)&\includegraphics[width=.9\columnwidth]{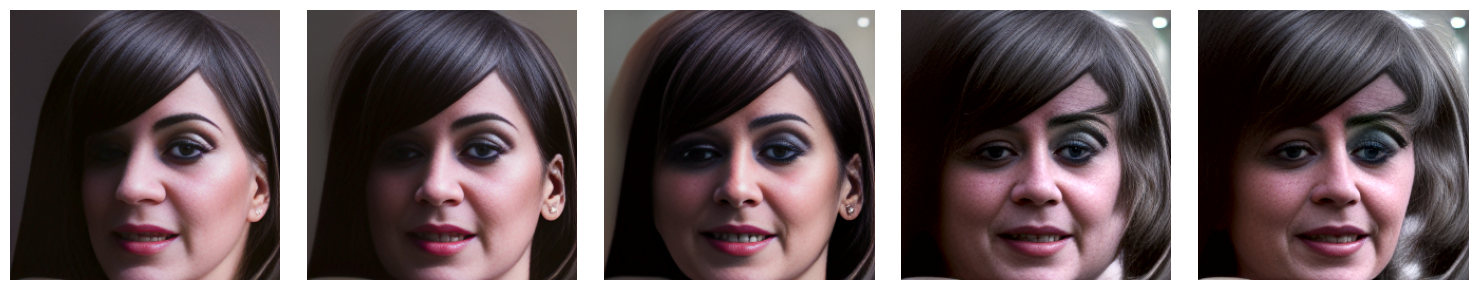}\\
\\
\end{tabular}
\caption{Relevance of attributes. For all images (a-d), the rotation in the first row corresponds to a trajectory computed just considering angles, while the second one is relative to a trajectory taking attributes into account. These are examples of complex rotations due to
the strong shadows over the face.\label{fig:relevance_of_attributes}}
\end{figure}

The relevance of exploiting attributes is exemplified in 
Figure~\ref{fig:relevance_of_attributes}, where we compare rotations obtained selecting clouds of images just according to rotations (first row), with the case where we refine the selection with relevant attributes (second row).

In Section~\ref{sec:slopes_comparison}, we provide a more technical comparison of the different trajectories in terms of their cosine similarity. We also used this metric as a way to select the most relevant attributes. There is a delicate balance between the specificity provided by attributes and the statistical relevance of the images retrieved from the dataset, which is essential for the regression phase. More details can be found in \cite{guerra2023exploring}. 

\section{Preprocessing and Postprocessing}
\label{sec:pre_post_processing}
The deployment of the previous technique requires a few preprocessing and postprocessing steps, discussed in this section. Preprocessing is aimed to prepare inputs in a format suitable for the DDIM embedder, while post-processing is devoted to enhance the quality of the result. 

\subsection{Preprocessing}
In this article, we restrict the input to aligned CelebA images. We could generalize the approach to an arbitrary image provided by the user, as we did in \cite{reification}, but the scientific added value is negligible.

Since the input image is already aligned, we work with a central crop
of dimension $128\times128$, frequently used in the literature \cite{TwoStage}, and then resized to dimension $64 \times 64$. 

The main step of the preprocessing phase is the background removal, since we experimentally observed that this operation facilitate rotation.
To this aim, we trained a U-Net model on the CelebAMask-HQ dataset, which includes high-quality, manually annotated face masks. All masks were combined and treated as a binary segmentation problem, focusing on background/foreground separation.

\begin{figure}[H]
\centering
\begin{tabular}{cc}
\includegraphics[width=\columnwidth]{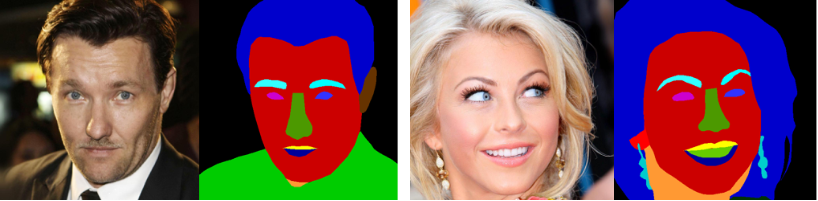} \\
(a) \\
\includegraphics[width=\columnwidth]{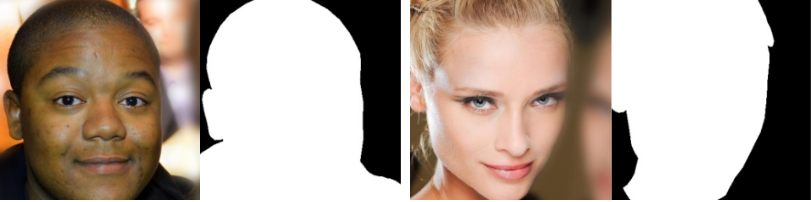}\\
(b)            
\end{tabular}
\caption{(a) examples of CelebaMask-HQ segmentations; (b) cropped version with unified masks used to train the model on the background removal task.}
\label{fig:segmentation_model_dataset}
\end{figure}
This approach allowed us to obtain a fairly precise segmentation of the facial region, with a precision of $96.78\%$ and a recall of $97.60\%$.

\subsection{Postprocessing}
To enhance the final results, we crafted a postprocessing pipeline featuring two additional steps: super-resolution and color correction. This meticulous process ensures sharper details and more accurate colors.

\subsubsection{Super-resolution}
The initial output, generated at a resolution of 64 × 64, is enhanced to 256 × 256 using CodeFormer \cite{codeformer}, a recently introduced model known for its proficiency in Super-Resolution and Blind Face Restoration. CodeFormer amalgamates the strengths of transformers and codebooks to achieve remarkable results. Transformers have gained large popularity and widespread application in natural language processing and computer vision tasks. On the other hand, codebooks serve as a method to quantize and represent data more efficiently in a compact form. The codebook is learned by self-reconstruction of HQ faces using a vector-quantized autoencoder, which embeds the image into a representation capturing the rich HQ details required for face restoration.

The key advantage of employing Codebook Lookup Transformers for face restoration lies in their ability to capture and exploit the structural and semantic characteristics of facial images. 
By employing a pre-defined codebook that encapsulates facial features, the model proficiently restores high-quality face images from low-quality or degraded inputs, effectively handling various types of noise, artifacts, and occlusions.

\subsubsection{Color Correction}
The final step of the post-processing phase involves applying a color correction technique to reduce color discrepancies between the generated faces and their corresponding source images. This technique is essential for enhancing the overall visual coherence of the final result.

The color correction process leverages the Lab color space to match the color statistics of the two images. It begins with converting both images to the Lab color space. Then, the Lab channels of the target image are adjusted by normalizing them according to the mean and standard deviation of the source image. Finally, the target image is converted back to the RGB color space, ensuring that the colors of the generated face closely match those of the original source image.

Once the trajectory is identified, we move along it for a specified number of steps, checking the rotation after each iteration. If the generated image does not show the expected rotation, we try to dynamically increase the number of steps.

In case of images with a large initial yaw, we also apply a preliminary face frontalization phase.

\section{Analysis of the Trajectory Slopes}
\label{sec:slopes_comparison}

In this section, we contrast the trajectory slopes within the latent space of Diffusion Models acquired through distinct attribute selections. We utilize cosine similarity as a synthetic metric to gauge the correlation between these trajectories.

We recall that we approximate trajectories using linear steps, derived from linear regression conducted over the centroids of diverse clusters of data point embeddings. The selection of data points is based on yaw angles and various attributes, comprising source of illumination.

\begin{figure}[H]
    \centering
    \includegraphics[width=0.7\columnwidth] {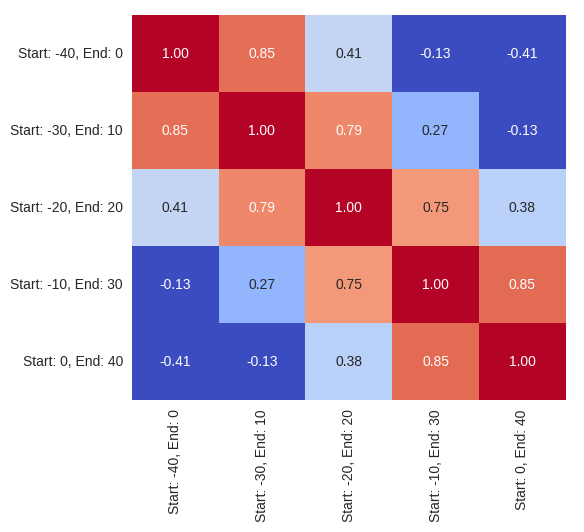}
    \caption{Cosine Similarities between trajectory slopes. The different trajectories are obtained from data whose rotation yaw is comprised in the specified range.}    \label{fig:cosine_similarity_differentStartingPoint}
\end{figure}

Figure \ref{fig:cosine_similarity_differentStartingPoint} provides a visual representation of the variation
of trajectory slopes across varying ranges of rotation degrees.  A heatmap is employed to graphically portray the level of resemblance between the trajectories, where distinct colors denote the magnitude of similarity.

As depicted in the figure, altering the degree ranges employed for cluster creation by 10 degrees leads to a consistently diminishing cosine similarity among the slopes. The most significant disparity is observable between the intervals [$-40^\circ$, $0^\circ$] and [$0^\circ$, $+40^\circ$]. This means that the direction of the trajectory required to turn a face in the range [$-40^\circ$, $0^\circ$] is very different from the direction required to turn it in the range [$0^\circ$, $+40^\circ$]. 

Based on the preceding analysis, it might appear that an incremental rotation approach involving frequent slope recalculations holds an advantage. Nevertheless, following the slope computation, it becomes necessary to translate the trajectory from centroids to the latent representation of the current image, and move from it. With each step, there is usually a gradual erosion of individual facial attributes. Thus, a trade-off arises: fewer steps could result in a relatively less precise rotation but better preservation of identity traits, while a greater number of steps could yield the opposite outcome.

According to our experimental findings, we obtained the best outcomes by just using two trajectories: the rightward trajectory [$0^\circ$, $-40^\circ$] and the leftward trajectory [$0^\circ$, $+40^\circ$]. This is
typically preceded by a frontalization step, when required.

The remainder of this section is dedicated to evaluating the influence of auxiliary attributes on trajectory definition: does rotating a male head yield the same results as rotating a female one? What implications arise from factors such as age or face illumination?

\begin{figure}[H]
    \centering
    \includegraphics[width=.8\columnwidth] {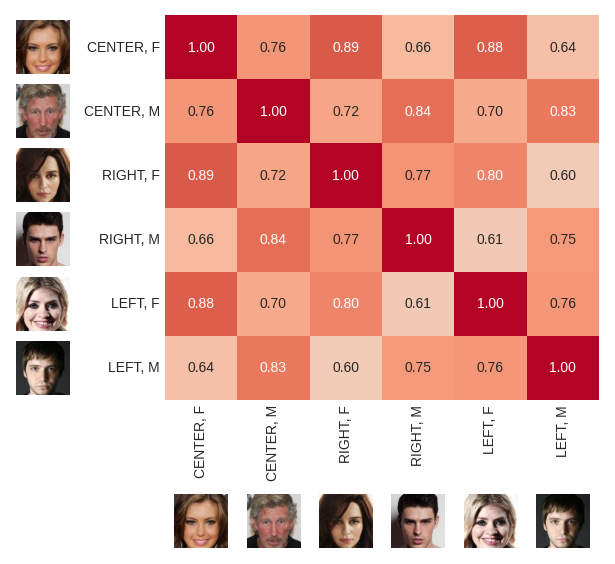}\\
    (a)\\
     \includegraphics[width=.8\columnwidth]{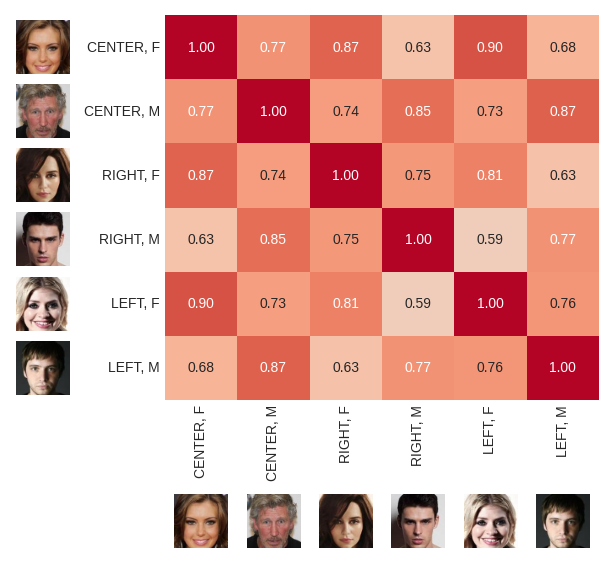}\\
    (b)
    \caption{Heatmap of Cosine Similarities Between Left (a) and Right (b) Slopes, computed using only  Light Direction (CENTER, LEFT or RIGHT) and Gender (M or F) as attributes. Each subset of attributes is represented by a corresponding sample from CelebA.}
    \label{fig:cosine_similarity_lightDirectionAndGender_slopes}
\end{figure}

\begin{figure}[H]
    \centering
    \includegraphics[width=.8\columnwidth] {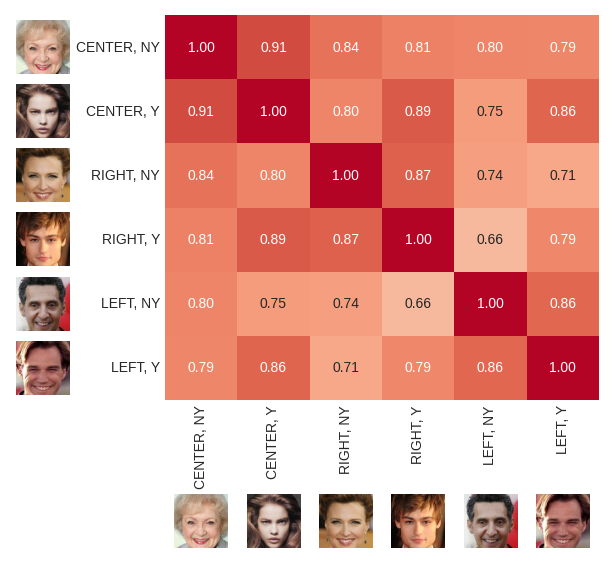}\\
    (a)\\
     \includegraphics[width=.8\columnwidth]{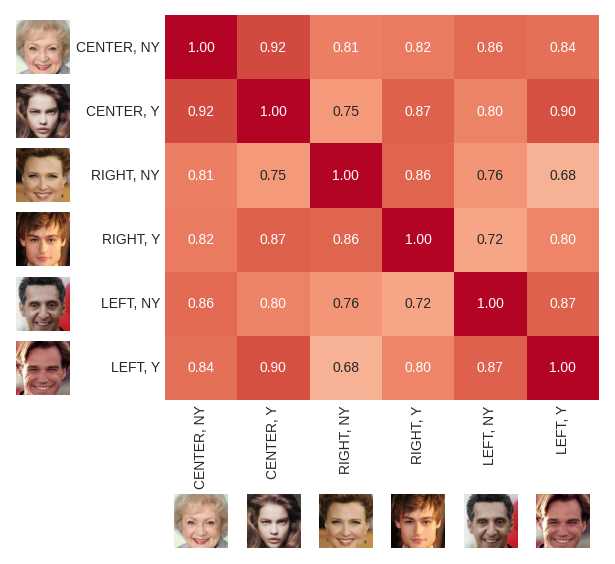}\\
     (b)
    \caption{Heatmap of Cosine Similarities Between Left (a) and Right (b) Slopes, computed using only  Light Direction (CENTER, LEFT or RIGHT) and Youthfulness (Y or NY) as attributes.}
    \label{fig:cosine_similarity_lightDirectionAndYouthfulness_slopes}
\end{figure}

\begin{figure}[H]
    \centering
    \includegraphics[width=.7\columnwidth] {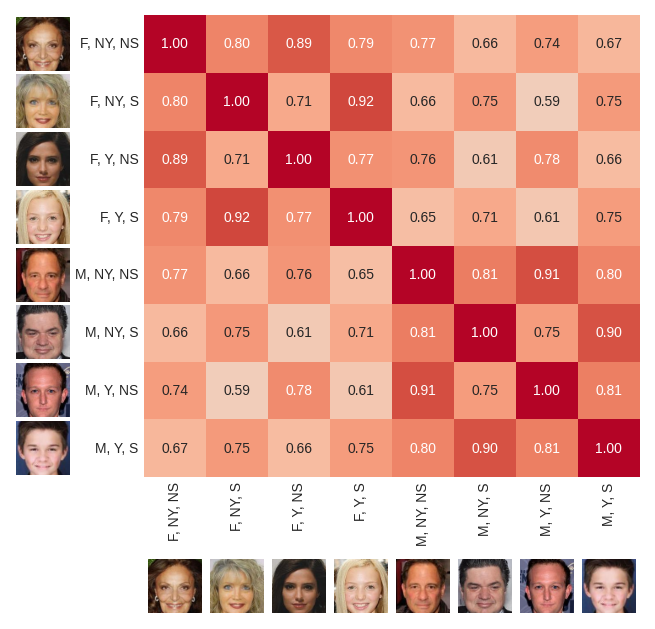}\\
    (a)\\
     \includegraphics[width=.7\columnwidth]{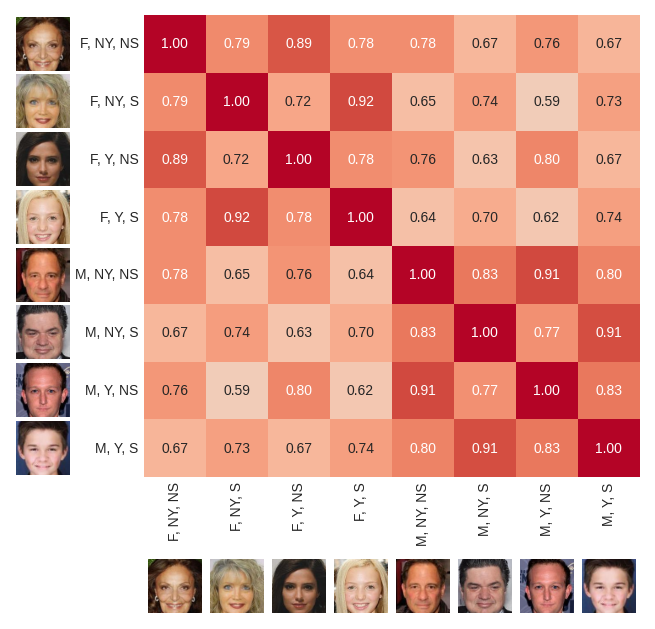}\\
     (b)
    \caption{Heatmap of Cosine Similarities Between Left (a) and Right (b) Slopes, computed with Light Direction fixed to center and using only Gender (M or F), Youthfulness (Y or NY), and Smiling (S or NS) as attributes.}
    \label{fig:cosine_similarity_AllAttributesCenterLight_slopes_withImages}
\end{figure}

To this aim, we fix an initial central pose,
two fixed trajectories, rightward and leftward,   
and compare the slopes obtained selecting
data points for centroids according to different
attributes. 
Specifically, in Figures \ref{fig:cosine_similarity_lightDirectionAndGender_slopes} we focus on gender and illumination, in 
Figure \ref{fig:cosine_similarity_lightDirectionAndYouthfulness_slopes} on illumination and age, and finally in Figure 
\ref{fig:cosine_similarity_AllAttributesCenterLight_slopes_withImages} on gender, age and smile. 
In all the Figures, the individual faces are merely 
meant to be a representative of their corresponding class of attributes, and have no specific influence
on the slope of the associated trajectory.

Figure \ref{fig:cosine_similarity_lightDirectionAndGender_slopes} presents a heatmap that illustrates the cosine similarities between the left and right slopes computed on clusters of images taking into account only two attributes: light direction (Center, Left, or Right) and gender (Male or Female). Gender is the most important factor of variation, more relevant than 
illumination source. Still, however, images with disparate light directions results in sensibly different trajectories.

These results suggest that both light direction and gender are important attributes for enhancing accuracy in trajectory calculation.

Figure \ref{fig:cosine_similarity_lightDirectionAndYouthfulness_slopes} parallels the previous concept, differing solely in the replacement of gender with age (Young or Not Young) as the attribute under consideration.

As depicted by the figure, a consistent trend emerges: diminished similarity values occur when the light direction is in contrast, and the same trend applies to the age parameter. Nevertheless, it's noteworthy that age has a relatively lesser impact on trajectory definition.

Figure \ref{fig:cosine_similarity_AllAttributesCenterLight_slopes_withImages} embarks on a more intricate exploration of trajectory attributes. 
In this context, we delve into a refined set of attributes: Gender (M or F), Youthfulness (Y or NY), and Smiling (S or NS). 

Similar to previous instances, the lowest similarity emerges when all attributes stand distinct from each other. For instance, a comparison between a female who is not young and not smiling ([F, NY, NS]) and a male who is young and not smiling ([M, Y, NS]) represents the scenario with the least similarity. Furthermore, upon closer examination of the image, it becomes apparent that gender exerts the most significant impact on similarity, closely trailed by the "smiling" attribute.

In our work, we consistently employed the aforementioned technique to methodically select the most pertinent attributes for delineating trajectories within the latent space.

\section{Results and troubleshooting}
\label{sec:results}
Measuring the quality of generative systems is a notoriously difficult task, due to the lack of a ground truth to compare with. This is particularly difficult in the case of the rotation operation, where we must assess both the model's capacity to obtain the desired orientation and the fidelity of the target to the source sample. Traditional metrics used in the field of generative modeling, like the Fréchet Inception Distance (FID), cannot be used in this context, since they are designed to compare distributions of data, not individual samples. In our case, the rotation measured on the generated sample is a parameter used to control the short iterative loop governing the computation of the trajectory; so, apart from a few cases where the algorithm fails to achieve the desired rotation and is forcibly stopped, the rotation of the result is the one expected.

The difficult task is to quantify the similarity of the individual features of the target with those
of the source. We are currently doing experiments with the Feature Similarity
Index (FSIM) \cite{FSIM}, the Identity Preservation Metric \cite{DeepFace}, 
ArcFace's Additive Angular Margin Loss \cite{ArcFace},
and the Learned Perceptual Image Patch Similarity (LPIPS) \cite{LPIPS}. All of them are valuable, but they also suffer from well-known limitations:
FSIM and similar metrics like SSIM are based on local patterns and luminance but may not adequately capture global context or the perceptual importance of different image regions; the Identity Preservation Metric heavily depends on the facial recognition or feature extraction model used, while the 
Learned Perceptual Image Patch Similarity (LPIPS) can be 
significantly influenced by the diversity and representativeness of the training dataset used for the neural networks that underpin this metric.
We shall report on these quantitative analyses in a forthcoming paper.

Also, a qualitative comparison with similar GAN-based architectures is problematic. As observed in \cite{comparingNCAA}, state-of-the-art GANs, especially when trained on CelebA-HQ, seem to have serious generative deficiencies: many images from CelebA seem to lie outside their generative range. This means that it is not always possible to embed a generic face in the latent space and reconstruct an image with sufficient similarity.

In this preliminary report, we shall just showcase the promising potential of our approach through some examples; the reader is also invited to test the system, freely available on GitHub at \href{https://github.com/asperti/Head-Rotation}{https://github.com/asperti/Head-Rotation}.

Some examples of rotations are given in Figure~\ref{fig:rotation_examples1}.
More examples are given in the appendix.
In general, the technique still suffers from of a few notable problems, and we shall devote the remaining part of this section to their discussion.

\begin{figure}[H]
    \centering
    \centering
      \includegraphics[width=.9\columnwidth]{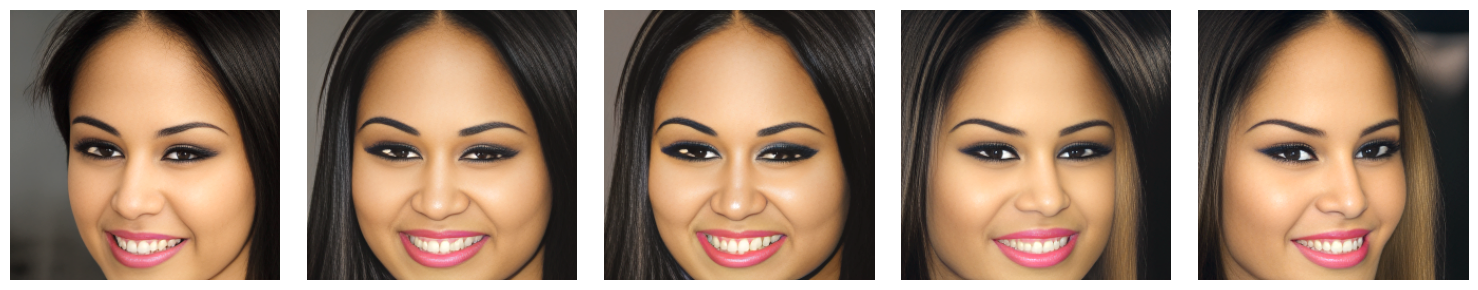}
      \includegraphics[width=.9\columnwidth]{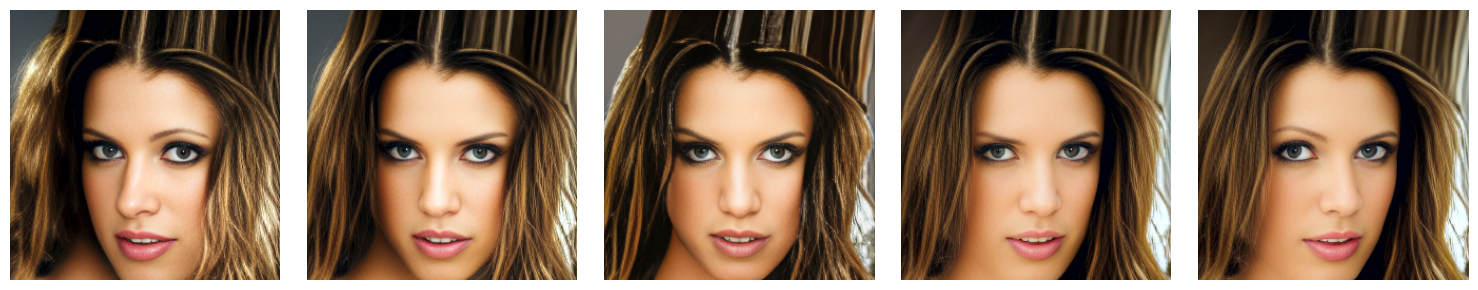}
      \includegraphics[width=.9\columnwidth]{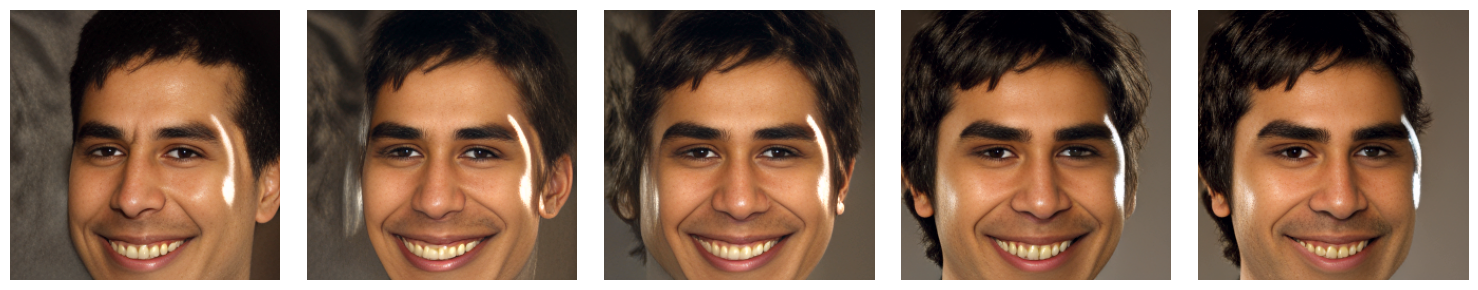}
      \includegraphics[width=.9\columnwidth]{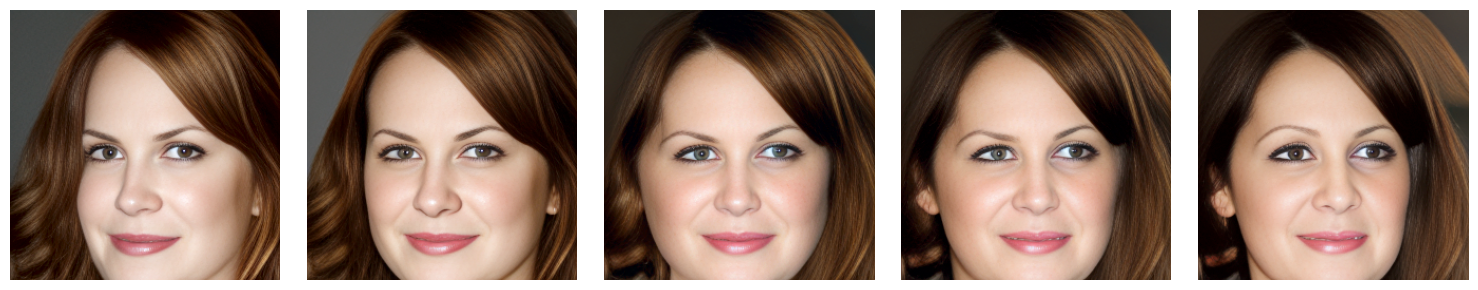}
       \includegraphics[width=.9\columnwidth]{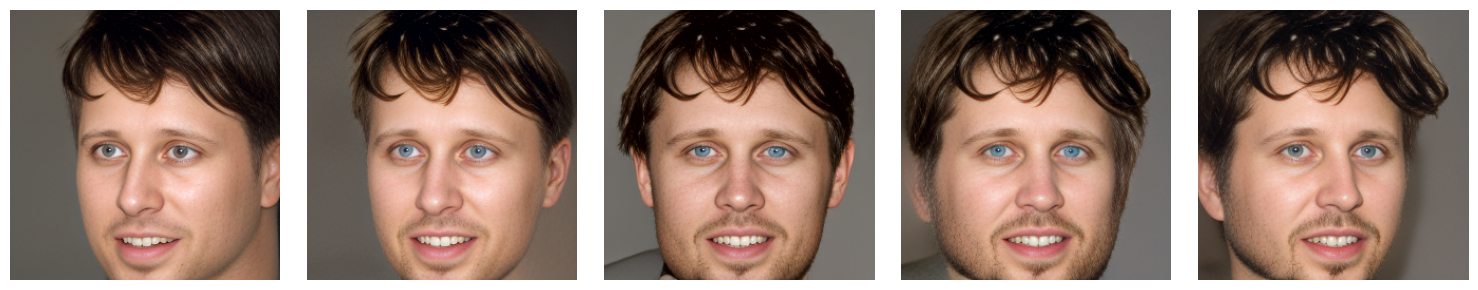}
        \includegraphics[width=.9\columnwidth]{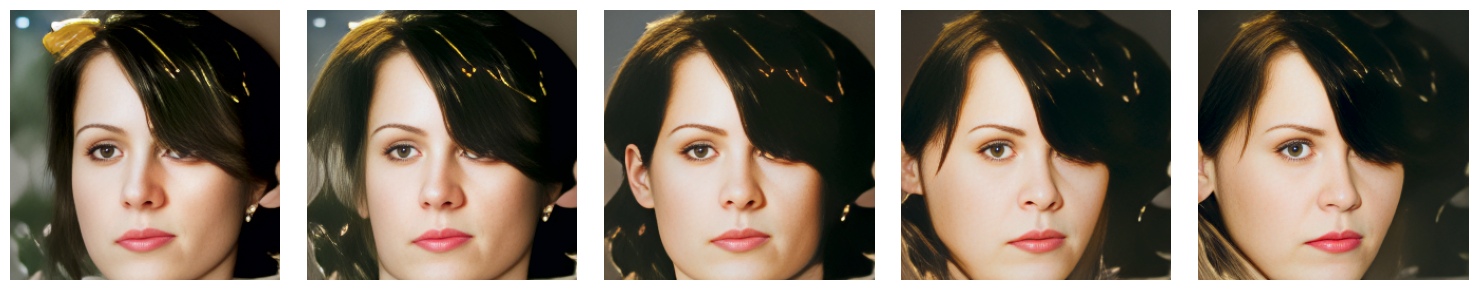}
        \includegraphics[width=.9\columnwidth]{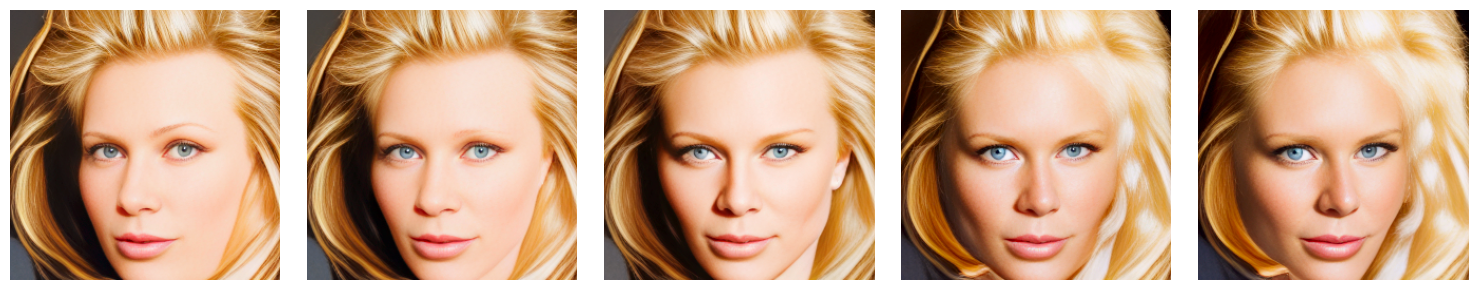}
        \includegraphics[width=.9\columnwidth]{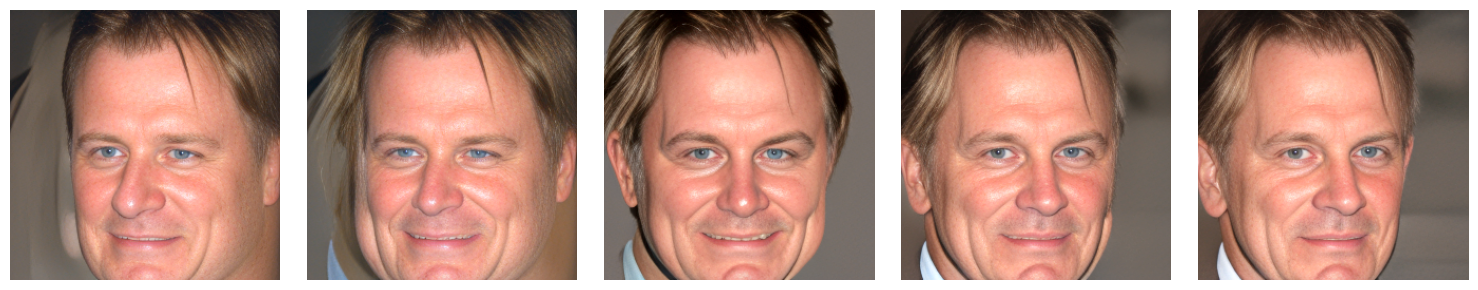}
        \includegraphics[width=.9\columnwidth]{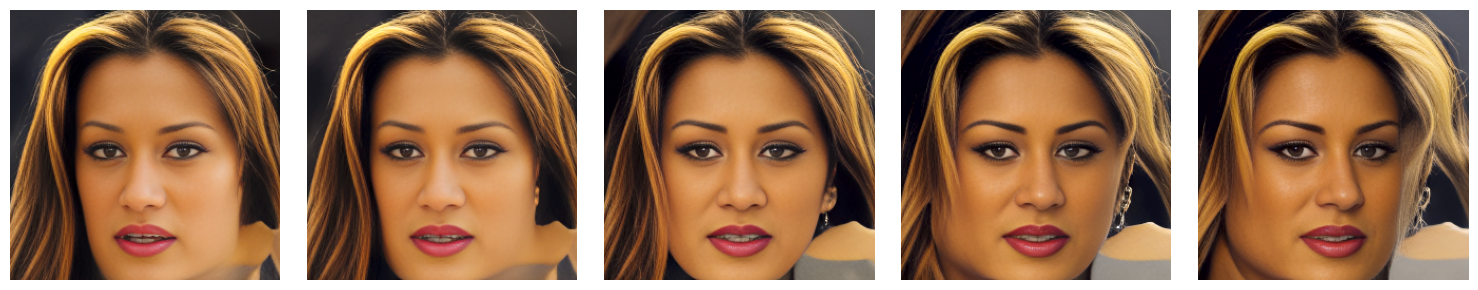}
    \caption{Examples of rotations. }
    \label{fig:rotation_examples1}
\end{figure}

\subsection{Loss of individual's features}
\begin{figure}[H]
    \centering
      \includegraphics[width=\columnwidth]{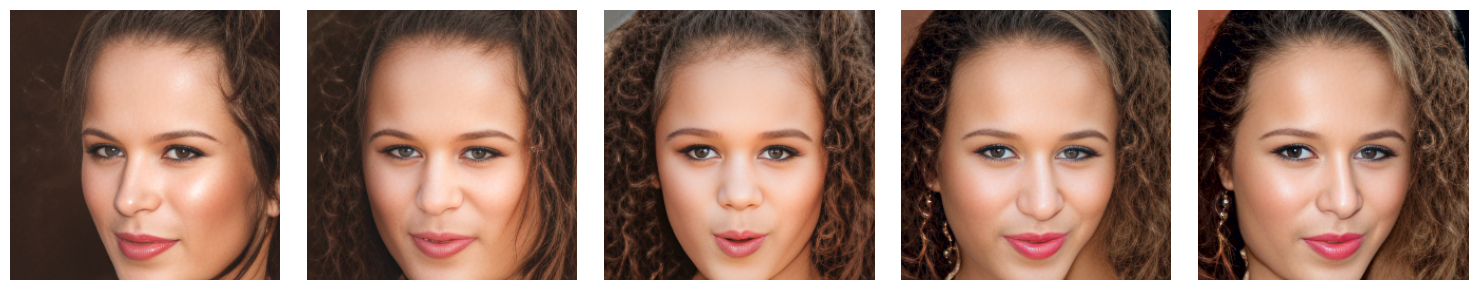}
    \caption{Troubleshooting: difficulty in preserving facial features}
    \label{fig:loss_of_identity}
\end{figure}

Preserving facial features while rotating the image of a large angle poses a significant challenge. This becomes especially problematic when employing a latent-based approach, where the essential traits of an individual are solely captured in the source point coordinates. Consequently, there's a risk of losing these key characteristics while following a given path, often resulting in more generic and less distinct features. Figure~\ref{fig:loss_of_identity} illustrates this phenomenon. While the right rotation (from the observer's point of view) appears reasonably accurate, the left rotation exhibits a gradual loss of the individual's features.

\subsection{High pitch and roll}
\begin{figure}[H]
    \centering
      \includegraphics[width=\columnwidth]{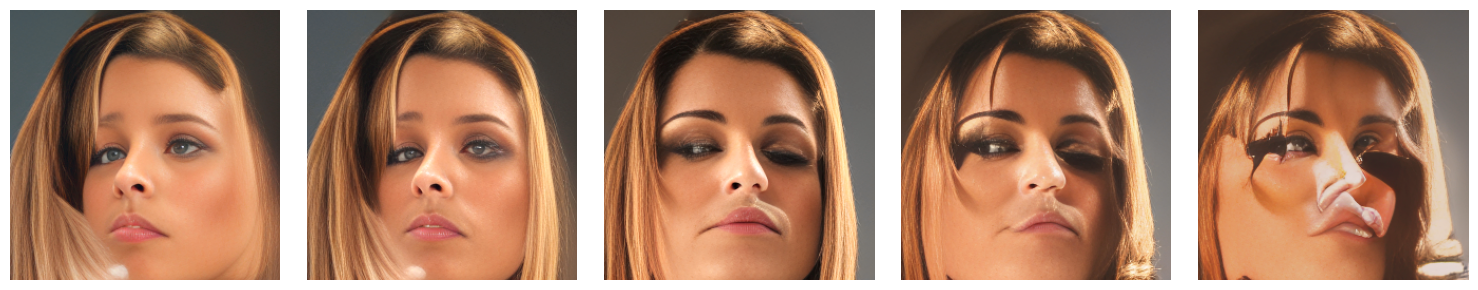}
    \caption{Troubleshooting: problems with high pitch or rolls}
    \label{fig:pitch_and_roll}
\end{figure}

In the presence of head poses with high pitch or roll (not
very frequent in CelebA), the technique can get into serious
troubles, as exemplified in Figure~\ref{fig:pitch_and_roll}

Usually, the method tries to either correct the anomalous angles, as along the left rotation, or simply gets lost, as along the right rotation.

\subsection{Hats and other artifacts}
\begin{figure}[H]
    \centering
      \includegraphics[width=\columnwidth]{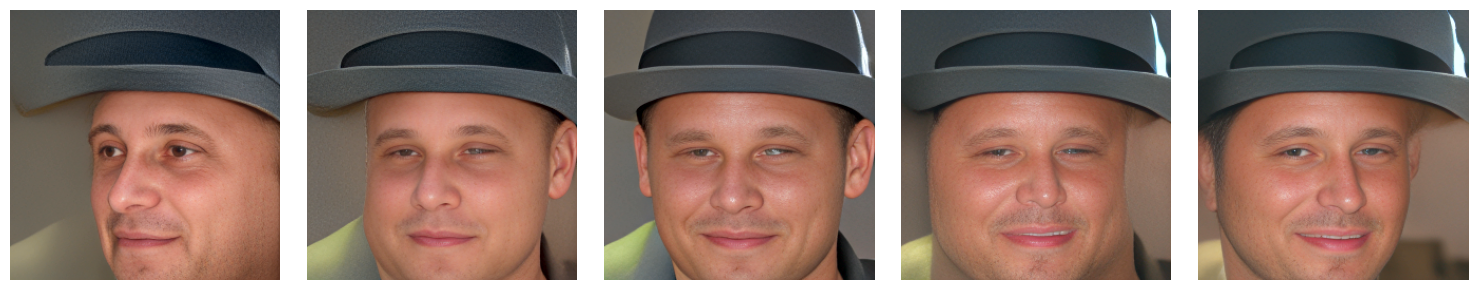}
      \includegraphics[width=\columnwidth]{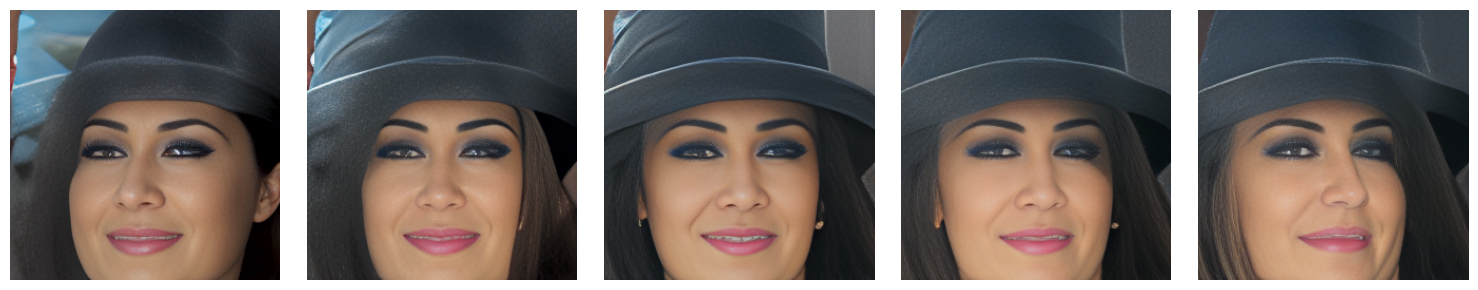}
    \caption{Troubleshooting: problems with Hats and other artifacts}
    \label{fig:hats}
\end{figure}

The generative model does not seem to have enough semantic information to handle situations involving the presence of
artifacts such as microphones, hats, or any kind of headgear
(see Figure~\ref{fig:hats}).

\begin{figure}[H]
    \centering
      \includegraphics[width=\columnwidth]{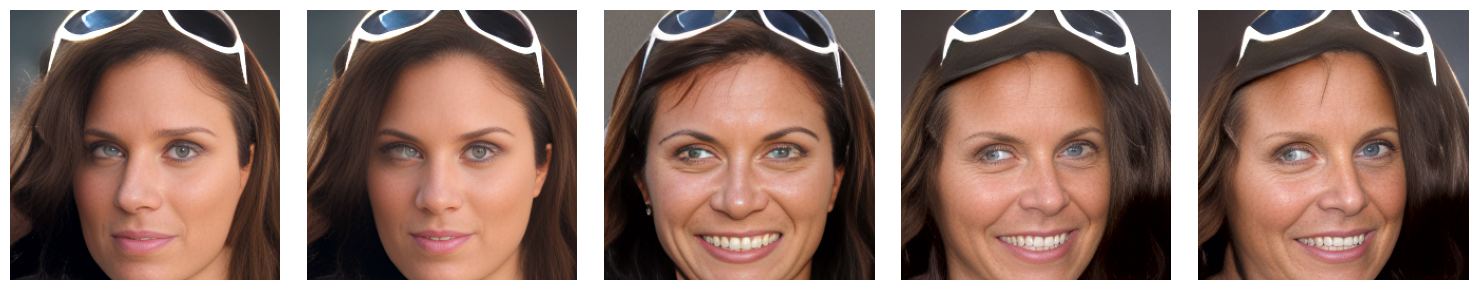}
    \caption{Troubleshooting: cannot rotate glasses over the head.}
    \label{fig:glasses_over_head}
\end{figure}

Also, glasses over the head are usually a problem, as exemplified in Figure ~\ref{fig:glasses_over_head}.
In the same figure, you may also observe the progressive loss of identity and change of expression along the left rotation.
Glasses over the eyes may be sometimes lost during rotation, but otherwise they are handled correctly. Several examples are given in 
the supplementary material.

\subsection{Deformation and loss of contours}
\begin{figure}[H]
    \centering
      \includegraphics[width=\columnwidth]{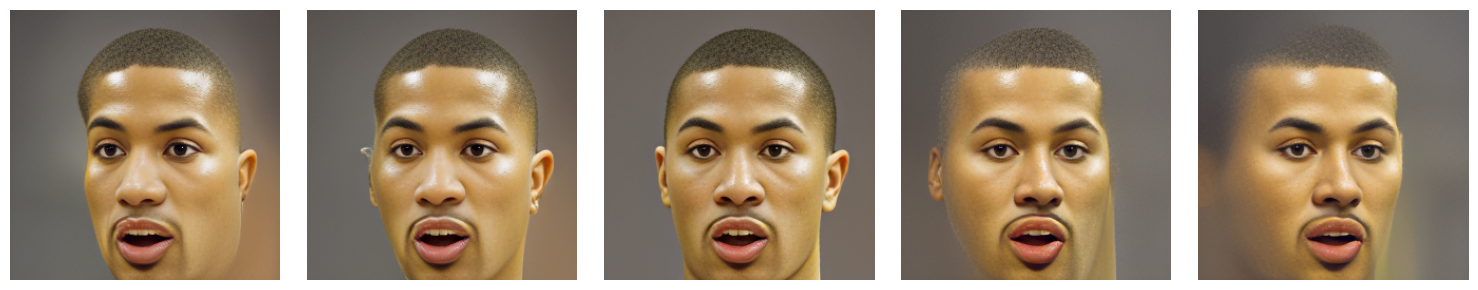}
    \caption{Troubleshooting: deformation and loss of contours.}
    \label{fig:loss_of_contours}
\end{figure}

Rotation may sometimes introduce anomalous deformations in the shape of the head; additionally, it is frequently unable to define
precise contours for the face under extreme yaw angles. Both 
phenomena are evident in Figure~\ref{fig:loss_of_contours}.

\subsection{Difficulty in rotating neck and ears}
\begin{figure}[H]
    \centering
     \includegraphics[width=\columnwidth]{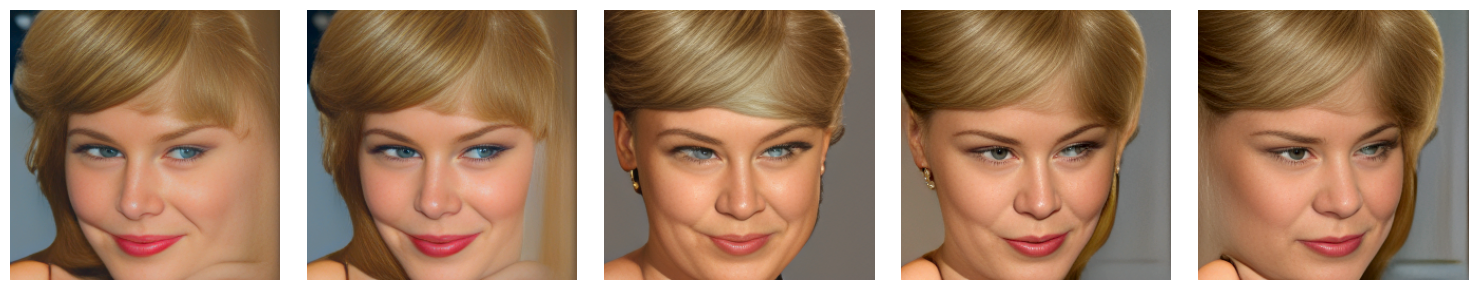}\\
      \includegraphics[width=\columnwidth]{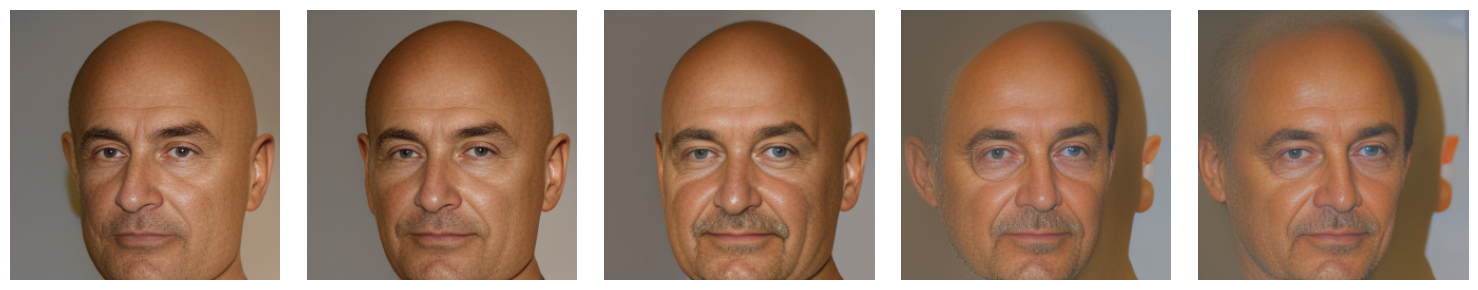}
    \caption{Troubleshooting: problems with neck and ears.}
    \label{fig:neck_and_ears}
\end{figure}

The technique is sometimes in trouble to correctly rotate the neck or ears of subjects. It may happen that they get 
detached from the actual figure, remaining in the ``background". 
This is illustrated in Figure~\ref{fig:neck_and_ears}.
Sometimes, a similar situation happens with hair, too.
 
\section{Conclusions}
\label{sec:conclusions}
This work contributes to the investigation of trajectories in the latent space of generative models, with particular attention to editing operations not easily expressible in terms of texture, color, or shapes of well-identified segmentation areas but requiring holistic manipulation of the image. Head rotation, especially intended as a continuous transformation, is a typical example of these kinds of manipulations. Our investigation suggests that identifying compelling trajectories relies on recognizing relevant attributes of the source image that can guide the statistical search in the latent space. Among these relevant attributes, in the case of head rotation, the direction of illumination plays a crucial role, creating complex shadowing effects on the face that are difficult to manage during rotation. Emphasizing the importance of lighting conditions for achieving realistic generative results in head rotation is one of the contributions of this work.

As a side result of our research, we created additional labels for the CelebA dataset, categorizing images into three groups based on the prevalent illumination direction: left, right, or center. Since CelebA-HQ is a well-known subset of CelebA, our labeling can be easily extended to the former dataset.

Our work is at a preliminary stage, and many aspects deserve further investigation.

Firstly, the current version lacks a robust quantitative evaluation and a thorough comparison with alternative techniques. Secondly, continuous movements should be better investigated in a video setting, a research field that has undergone remarkable achievements in recent years, mostly thanks to stable diffusion techniques 
\cite{VDM,ImagenVideo,sora2024}. Specifically, exploiting the spatio-temporal coherence of adjacent frames could help in understanding the global structure and 3D perspective, which becomes particularly useful when dealing with artifacts such as hats, earrings, or eyeglasses.

In the context of video generation, our work could contribute to extracting a dataset of difficult cases, especially in terms of light conditions, that could pose interesting challenges for generative models. This dataset would be valuable for testing and improving the robustness of generative models in handling complex scenarios, thereby advancing the field of video-based generative modeling.

Furthermore, our preliminary findings indicate that incorporating detailed lighting information into the generative process significantly enhances the realism of generated images. Future work should focus on developing more sophisticated methods for capturing and utilizing lighting attributes in the latent space. This includes exploring the use of advanced neural network architectures and loss functions specifically designed to preserve lighting consistency during image manipulation.

Additionally, we plan to extend our investigation to other types of holistic image manipulations beyond head rotation, such as changing facial expressions or body poses, which also require careful consideration of lighting and other contextual factors. By addressing these challenges, we contribute to provide a comprehensive framework for holistic image manipulation in generative models.


\funding{This research was partially funded by the Future AI Research (FAIR) project of the National Recovery and Resilience Plan (NRRP), Mission 4 Component 2 Investment 1.3 funded from the European Union - NextGenerationEU.}

\dataavailability{The application described in this paper is open
source. The software can be downloaded from the following github repository: \href{https://github.com/asperti/Head-Rotation}{https://github.com/asperti/Head-Rotation}}

\acknowledgments{We would like to thank the many students who helped in the annotation of CelebA for illumination orientation, and in particular L.~Bugo, D.~Filippini and A.~Rossolino. 
}

\conflictsofinterest{The authors declare no conflict of interest.}

\appendixtitles{yes} 
\appendixstart
\appendix

\section[\appendixname~\thesection]{Additional rotation examples}
In this appendix we provide a short list of additional examples of
rotations obtained by means of our model.


\begin{figure}[H]
    \centering
     \includegraphics[width=\textwidth]{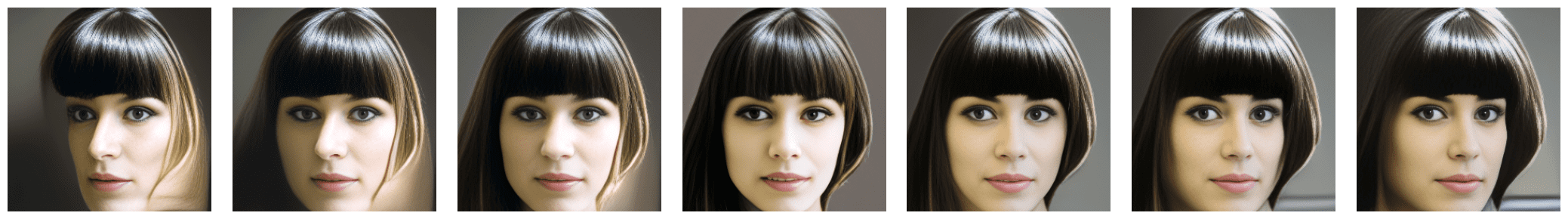}\\
     \includegraphics[width=\textwidth]{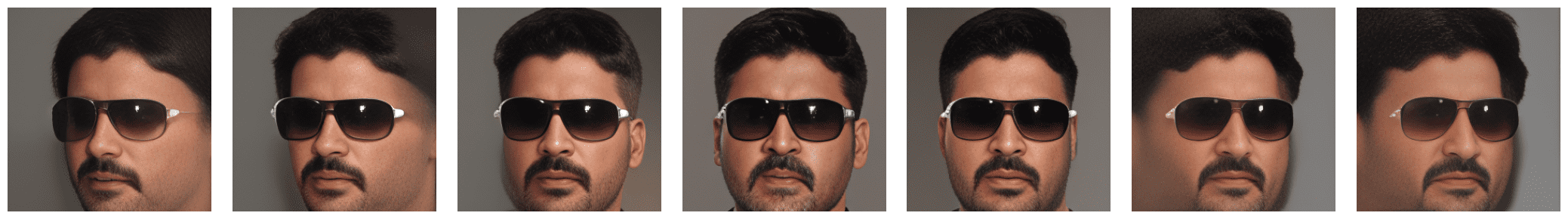}\\
     \includegraphics[width=\textwidth]{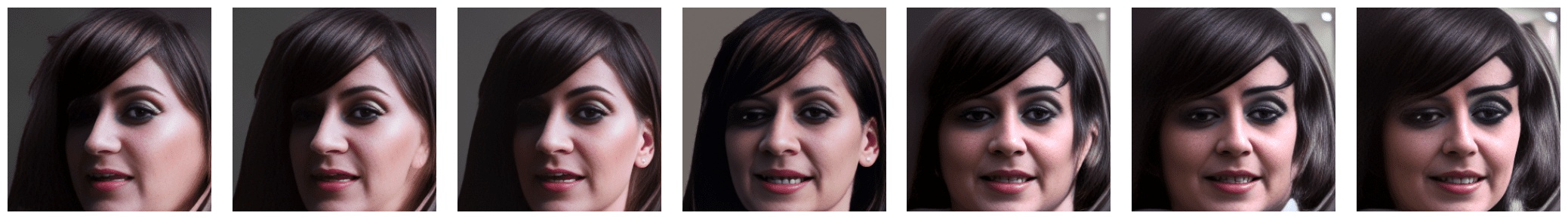}\\
      \includegraphics[width=\textwidth]{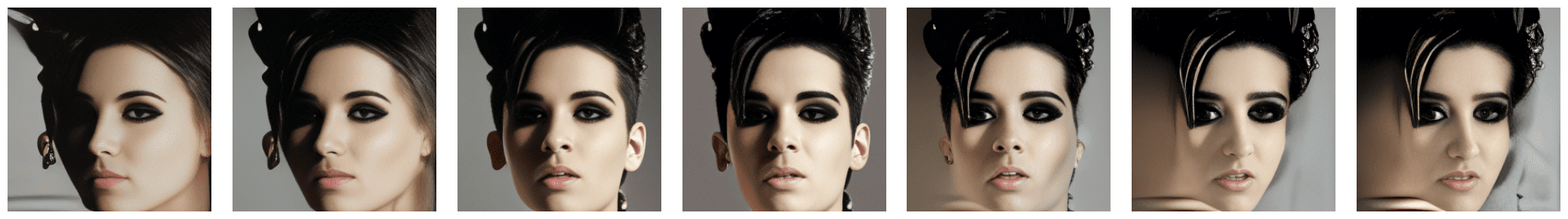}\\
    \includegraphics[width=\textwidth]{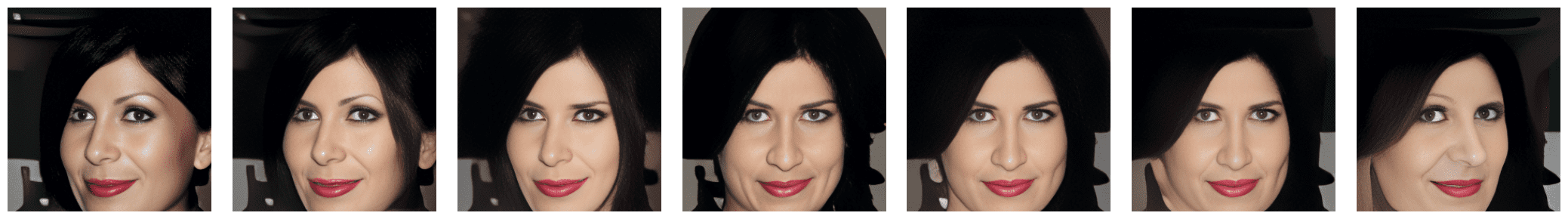}\\
     \includegraphics[width=\textwidth]{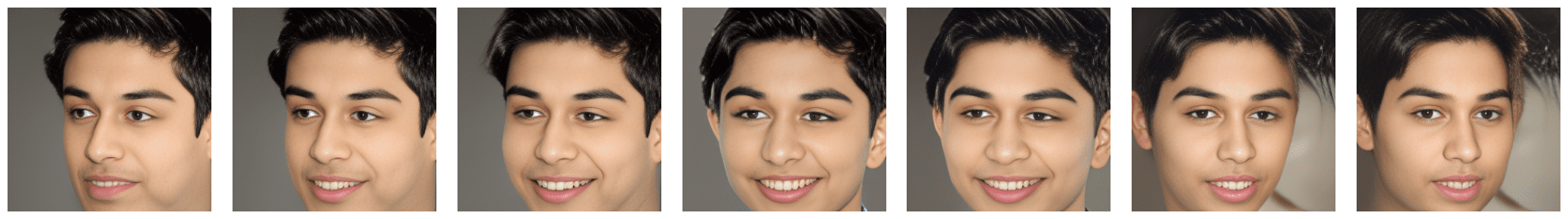}\\
    \includegraphics[width=\textwidth]{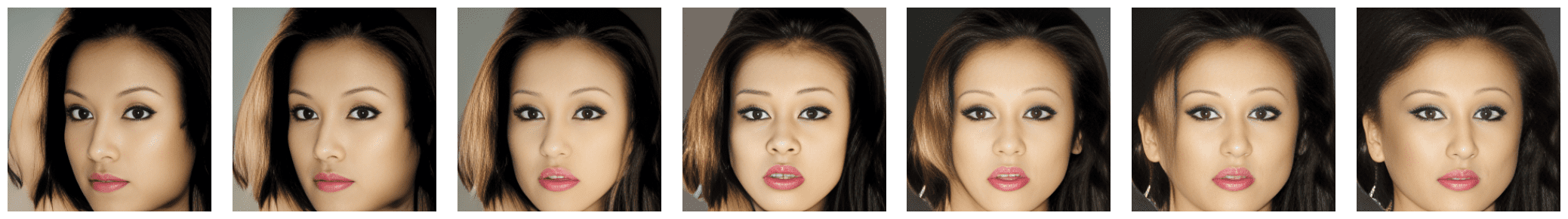}\\
      \includegraphics[width=\textwidth]{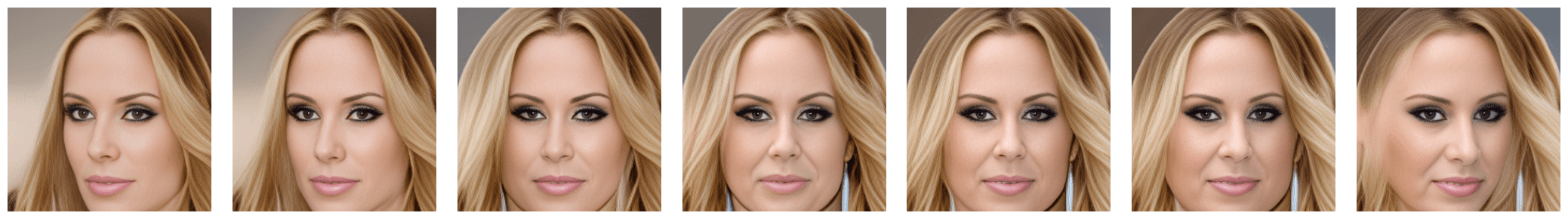}\\
      \includegraphics[width=\textwidth]{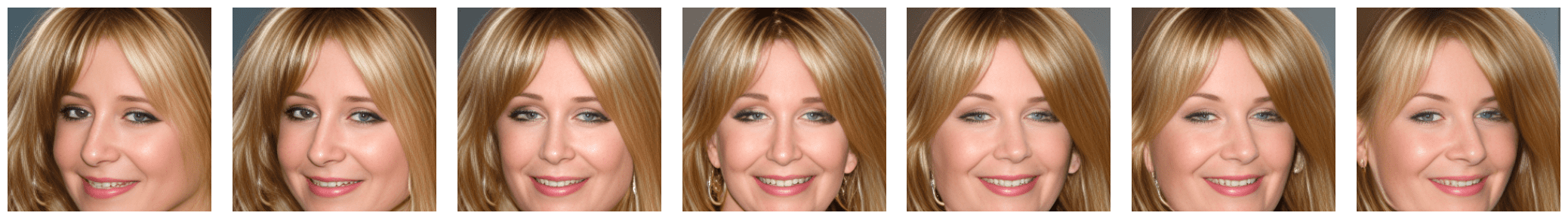}\\
      \includegraphics[width=\textwidth]{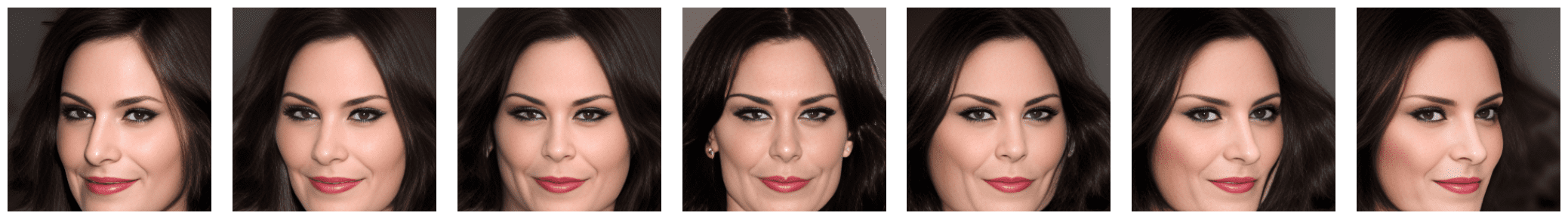}\\
      \includegraphics[width=\textwidth]{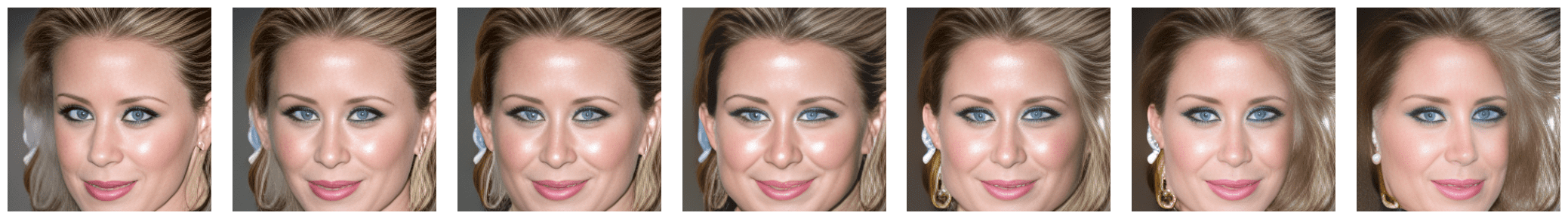}
    \caption{Examples of rotations.}
    \label{fig:more_examples1}
\end{figure}

\begin{figure}[H]
    \centering
       \includegraphics[width=\textwidth]{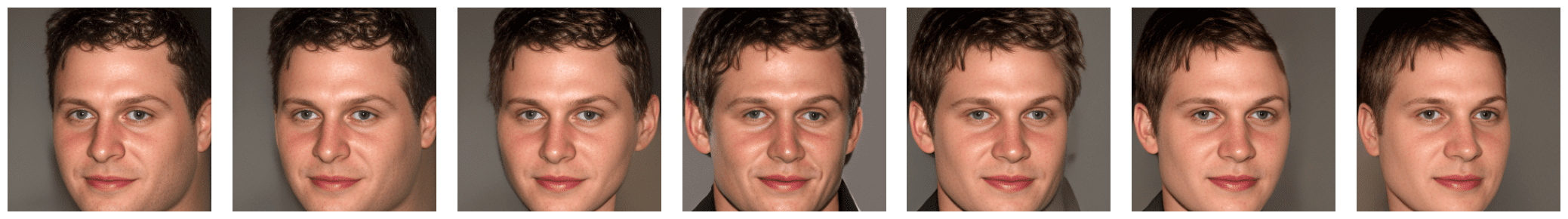}\\
      \includegraphics[width=\textwidth]{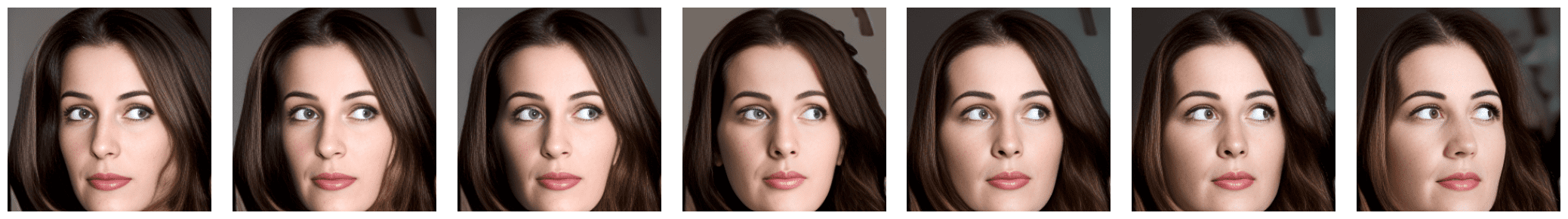}\\
      \includegraphics[width=\textwidth]{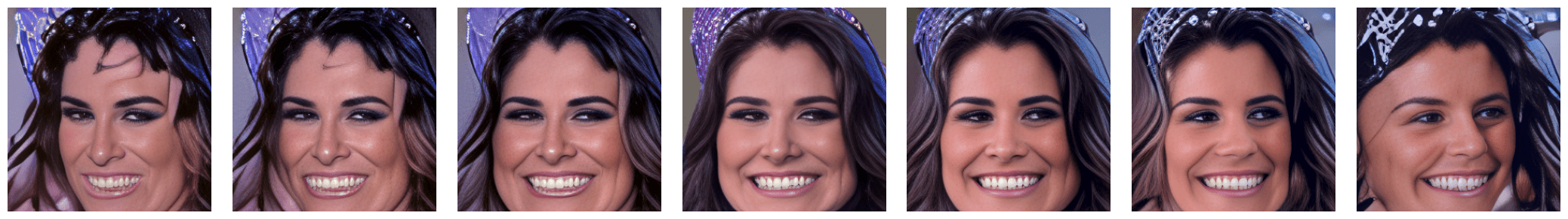}\\
      \includegraphics[width=\textwidth]{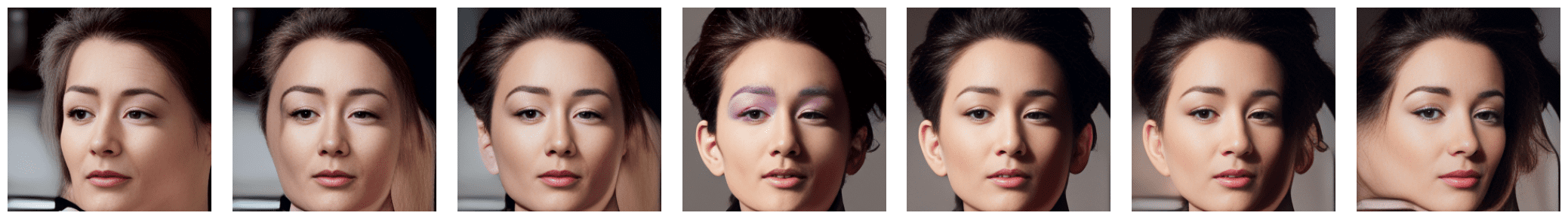}\\
       \includegraphics[width=\textwidth]{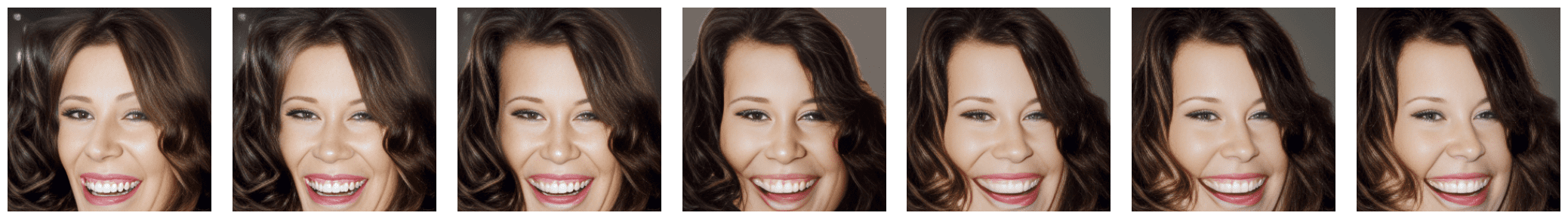}\\
      \includegraphics[width=\textwidth]{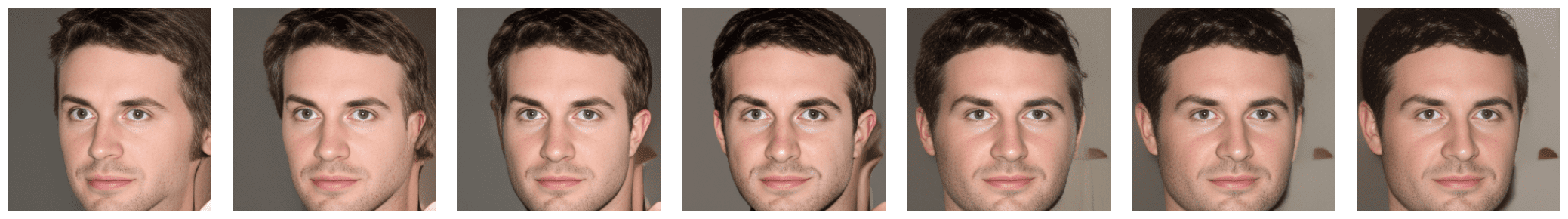}\\
       \includegraphics[width=\textwidth]{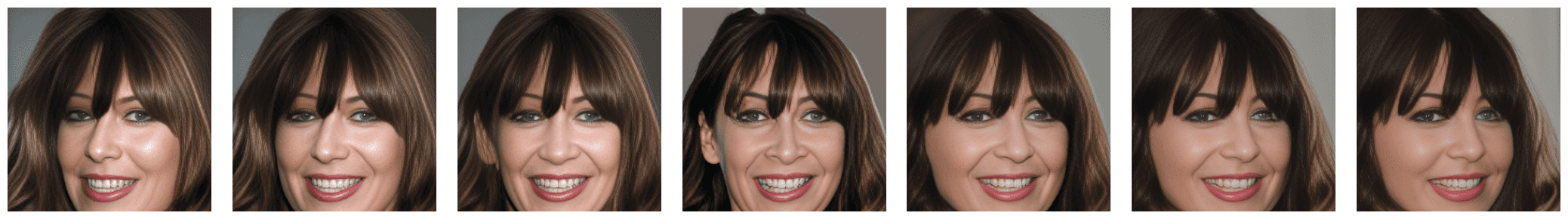}\\
      \includegraphics[width=\textwidth]{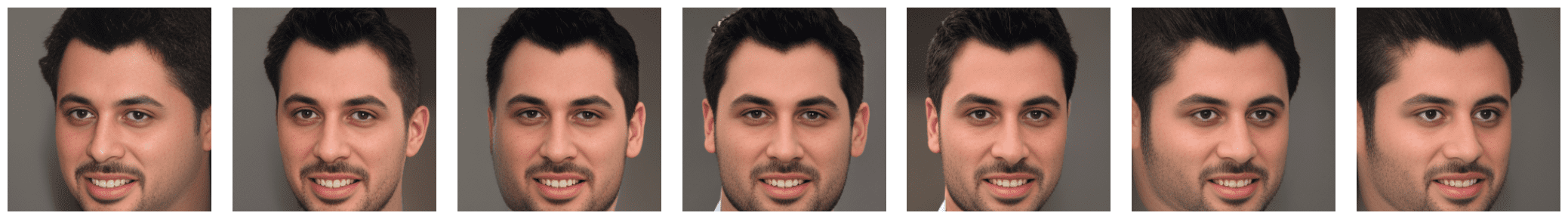}
    \caption{Examples of rotations.}
    \label{fig:enter-label}
\end{figure}

\begin{adjustwidth}{-\extralength}{0cm}

\reftitle{References}

\PublishersNote{}
\end{adjustwidth}
\end{document}